\documentclass[10pt,twocolumn,letterpaper]{article}

\usepackage{iccv}       
\usepackage{multirow}

\definecolor{iccvblue}{rgb}{0.21,0.49,0.74}
\usepackage[pagebackref,breaklinks,colorlinks,allcolors=iccvblue]{hyperref}

\definecolor{CQColor}{rgb}{0.0,0.0,1.0} 

\definecolor{TSColor}{rgb}{0.5,0.0,0.8} 

\definecolor{CQRColor}{rgb}{1.0,0.0,1.0} 

\usepackage{bbding}

\definecolor{leakage}{HTML}{FF0000}
\definecolor{artifacts}{HTML}{00CC00}

\def\paperID{7197} 
\def\confName{ICCV}
\def\confYear{2025}

\title{FixTalk: Taming Identity Leakage for High-Quality \\Talking Head Generation in Extreme Cases}

\author{
Shuai Tan, Bill Gong, Bin Ji, and Ye Pan\thanks{Corresponding author.} \\
Shanghai Jiao Tong University \\
{\tt\small \{tanshuai0219\}@sjtu.edu.cn}
}

\begin{document}

\twocolumn[{
\renewcommand\twocolumn[1][]{#1}
\maketitle
\begin{center}
    \captionsetup{type=figure}
    \includegraphics[width=0.96\textwidth]{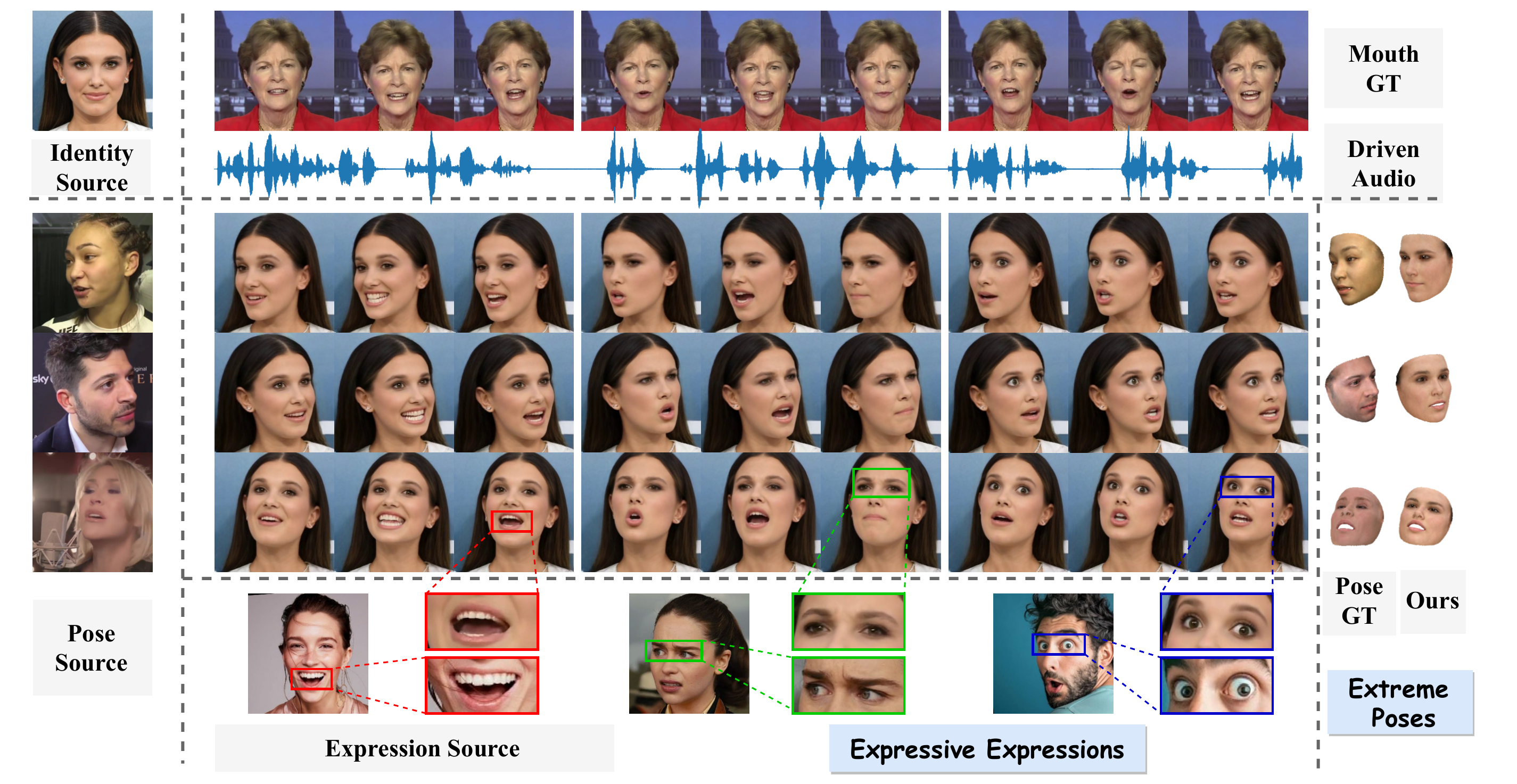}
    \captionof{figure}{\textbf{Example animations generated by FixTalk.} Given an identity source, FixTalk supports \textit{decoupled control} over lip sync, head pose, and emotional expressions from their respective sources. As presented in~\cref{tab:efficiency}, FixTalk achieves \textit{real-time generation}. In particular, FixTalk demonstrates superior performance with \textit{high-quality rendering} in extreme poses and expressive expressions. }
    \label{fig:teaser}
\end{center}
}]

\begin{abstract}
Talking head generation is gaining significant importance across various domains, with a growing demand for high-quality rendering. However, existing methods often suffer from identity leakage (IL) and rendering artifacts (RA), particularly in extreme cases. Through an in-depth analysis of previous approaches, we identify two key insights: (1) IL arises from identity information embedded within motion features, and (2) this identity information can be leveraged to address RA. Building on these findings, this paper introduces FixTalk, a novel framework designed to simultaneously resolve both issues for high-quality talking head generation. Firstly, we propose an \textbf{Enhanced Motion Indicator (EMI)} to effectively decouple identity information from motion features, mitigating the impact of IL on generated talking heads. To address RA, we introduce an \textbf{Enhanced Detail Indicator (EDI)}, which utilizes the leaked identity information to supplement missing details, thus fixing the artifacts. Extensive experiments demonstrate that FixTalk effectively mitigates IL and RA, achieving superior performance compared to state-of-the-art methods.
\end{abstract}
    
\section{Introduction}
\label{sec:intro}

Talking head generation~\cite{chen2019hierarchical, jamaludin2019you, wang2021audio2head, zhou2019talking}, the task of producing realistic facial animations driven by audio or other inputs, has garnered significant interest due to its wide range of applications in areas like virtual digital humans, filmmaking, and education. To facilitate the above application, modern talking head generation methods are expected to achieve three critical goals: (1) \textit{System Efficiency}: Given the potential real-time applications, 
fast inference and minimal computational overhead are necessary for broader adoption~\cite{guo2024liveportrait}. (2) \textit{Decoupled Control}: Effective talking head generation involves finely controlling several key components, such as the mouth, head pose, and emotional expressions. To achieve realistic and expressive facial animations, it is crucial to disentangle and control these aspects independently, allowing for granular adjustments to each component without influencing the others~\cite{pd-fgc}. (3) \textit{High-Quality Rendering}: Producing high-quality video outputs is a primary goal for talking head generation, especially when creating extreme poses and expressive expressions~\cite{xie2024x}.

\begin{figure}[t]
  \centering
    \includegraphics[width=1\linewidth]{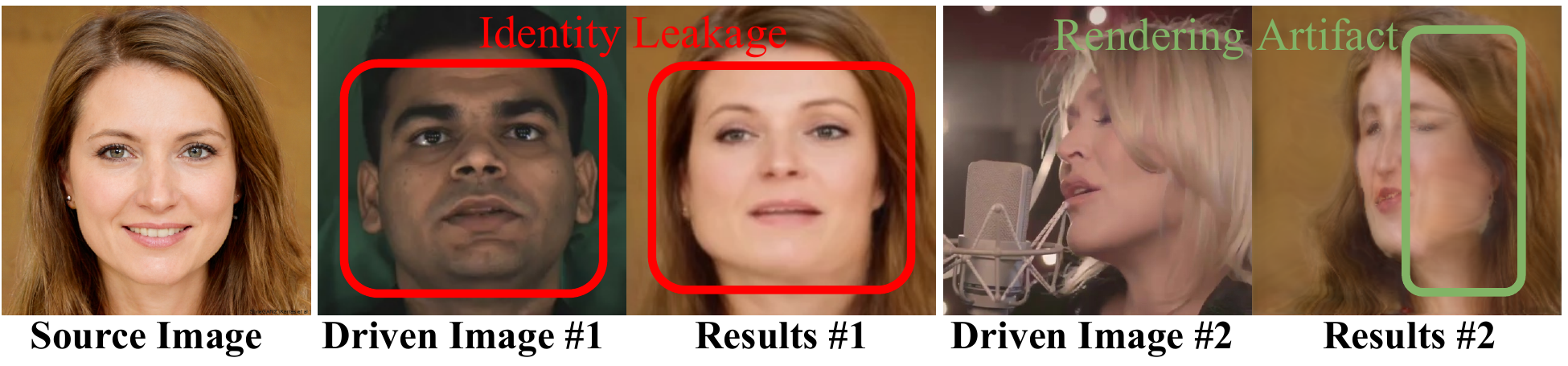}
    \caption{Illustration of identity leakage and rendering artifact.}
    \label{fig:limitation_introduction}
\end{figure}

To achieve the first two objectives, the model design is subject to numerous constraints, such as avoiding more effective diffusion frameworks~\cite{ma2024followyouremoji, xie2024x, xu2024hallo, mimir,tan2025SynMotion} to maintain high efficiency and decoupled controllability. This forces us to choose GAN-based~\cite{gan} models and face their potential issues related to low-quality rendering, which we refer to as \textbf{Identity Leakage} and \textbf{Rendering Artifacts}. 
Specifically, we test the state-of-the-art GAN-based talking head model~\cite{edtalk}, which has achieved real-time facial synthesis and disentangled facial control, revealing these issues manifesting as: the driven image's identity leaked to source image and obvious artifacts caused by detail loss in extreme pose shown in~\cref{fig:limitation_introduction}.
Therefore, we aim to address these two significant issues—commonly observed in GAN-based models for high-quality rendering—that notably hinder the widespread application of talking head generation.

\begin{figure*}[t]
  \centering
  \includegraphics[width=1\linewidth]{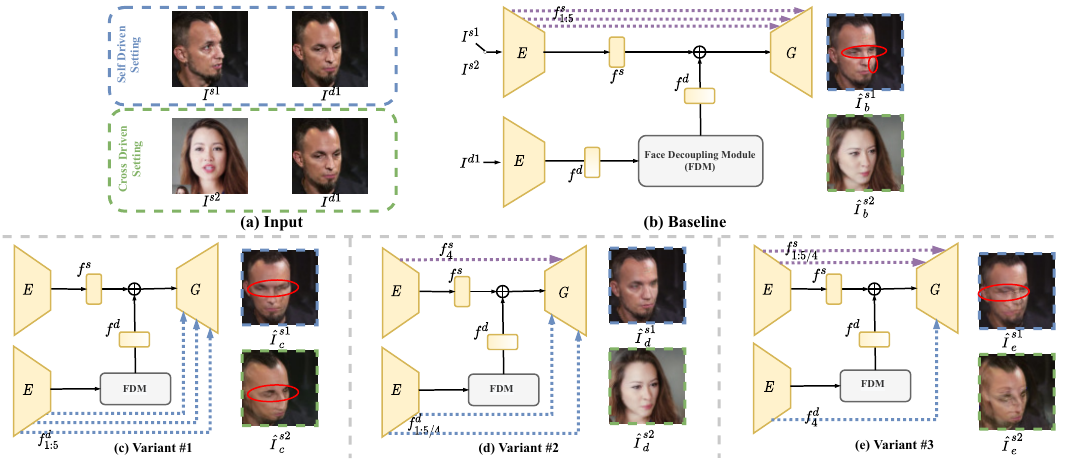}
    \caption{Latent feature investigation of Baseline.  (b) presents the overview of Baseline. The source image $I^{s}$ and the driven image $I^{d}$ are fed into an encoder to get the multi-scale features \{$f^s_{1:5}$, $f^s$\} and \{$f^d_{1:5}$, $f^d$\}. $f^d$ is further passed through FDM to get motion feature $f^{d}$, which is added to $f^s_{s}$ and fed into generator $G$. $G$ additionally incorporates $f^s_{1:5}$ for more identity-preservation results generation. (c)\&(d)\&(e): An exploration of the impact of different features from driven image on performance. }
    \label{fig:edtalk_limitation}
\end{figure*}

For this purpose, we investigate the causes of identity leakage in~\cref{sec:edtalk_limitation}. We find that (1) this leakage occurs primarily because the motion representation extracted from the driven image still retains identity information. Subsequently, we systematically explore which variable in the motion extraction process contains this identity information by incrementally replacing the intermediate variables of the driven image with those of the source image. This approach allows us to identify the specific identity-carrying variable within the motion representation. Interestingly, as shown in~\cref{fig:edtalk_limitation} (e), we observe that (2) the leaked identity information can be effectively utilized in a self-driven context to supplement the missing details in rendering artifacts, thereby improving rendering quality. These findings inspire us to tame identity leakage for high-quality rendering.

Building on these observations, in this paper, we introduce \textbf{FixTalk}, a new framework that addresses the limitations of {identity leakage} and {rendering artifacts} by leveraging the \textbf{Enhanced Motion Indicator (EMI)} and \textbf{Enhanced Detail Indicator (EDI)}.
Specifically, EMI first isolates motion-related features from the identity-carrying variable and utilizes multi-scale features to derive the motion representation, then leverages a new disentanglement loss function to achieve enhanced motion modeling.
Moreover, inspired by the beneficial identity leakage in self-driven setting, EDI extends this benefit to cross-driven animation by utilizing extra storage space to retain features that contribute to identity leakage during training. 
Subsequently, we design appropriate queries during inference to recall the most relevant identity-leaked features, assisting in eliminating artifacts related to extreme poses and exaggerated expressions. 

Notably, both EMI and EDI are lightweight and plug-and-play, widely adaptable to previous GAN-based works~\cite{lia, dpe, liu2024anitalker, liu2024vqtalker}. 
As illustrated in the example of~\cref{fig:teaser}, our FixTalk builds upon the popular base model EDTalk~\cite{edtalk}, preserving its advantages of high system efficiency and decoupled control while fixing its limitations in high-quality rendering (\textit{i.e.}, identity leakage and rendering artifacts), tackling the three key challenges of modern talking head generation.
%
%
Our main contributions are as follows: (1) We present FixTalk, which facilitates talking head generation to achieve three critical goals. (2) The investigation inspires us to propose EMI and EDI, which tame the identity leakage for high-quality talking head generation. (3) Extensive experiments demonstrate the superiority of FixTalk on the current talking head benchmarks.

\section{Related Work}
\label{sec:related_work}

\subsection{Video-Driven Image Animation}
Video-Driven Image Animation seeks to produce realistic face videos by maintaining the identity of a source image while replicating the motion from a driving video. On the one hand, benefiting from the better computational efficiency and controllability, the majority of animation approaches~\cite{megaportraits, yang2024smgdiff, zhao2024media2face, ma2024cvthead, zhang2023metaportrait, zakharov2019few, zhao2022thin, zhao2021sparse, tran2018nonlinear, wu2021f3a, zhu2017face, yao2020mesh, liang2024intergen} leverage generative adversarial networks (GANs) to learn motion dynamics to warp and render the final results. 
Wang \textit{et al.}~\cite{lia} generate animated videos by learning orthogonal motion directions within the latent space, which are linearly combined to avoid complex structural processing. Despite notable progress, these approaches still face limitations in identity leakage~\cite{liu2024anitalker}, and struggle to capture subtle or extreme out-of-domain facial expressions and head poses~\cite{ma2024followyouremoji, xie2024x}. Recently, sparked by the superior generative performance of the diffusion model~\cite{stablediffusion}, several works~\cite{xie2024x, ma2024followyouremoji, yang2024megactor, chen2024anifacediff, tan2024svp, wang2024unianimate, mimicmotion2024, musepose, peng2024controlnext, Animateanyone, magicdance, disco, magicanimate, zhu2024champ} have applied it into animation task and achieve impressive high-fidelity results through iterative refining process. X-portrait~\cite{xie2024x} introduces a local control module that emphasizes detailed facial expressions, crucial for conveying emotion but often challenging to capture. While these methods have achieved remarkable results in quality, they are computationally intensive and lack precise controllability~\cite{guo2024liveportrait}.

\subsection{Audio-Driven Talking Head Generation}
Audio-driven talking head generation methods~\cite{resyncer, guan2023stylesync, liu2024anitalker, lin2024cyberhost, loopy, xu2024hallo, mimaface, vexpress, aniportrait, echomimic, emo, xu2024vasa, cui2024hallo2, cui2024hallo3, echomimicv2,tan2024style2talker, ji2023stylevr, ji2025sport, tan2024animate,shi2024motionstone,tan2023emmn, tan2024flowvqtalker, tan2024say, pan2023emotional,pan2024expressive, pan2025vasa} are typically built upon video-driven image animation techniques, extending the task from a single video-driven modality to a cross-modal problem by incorporating audio input.
Some methods introduce intermediate representations as a bridge, leading to the design of two-stage GAN-based approaches~\cite{zhou2020makelttalk,chen2019hierarchical,das2020speech,zakharov2019few,zhong2023identity,wang2021audio2head,chen2020talking,yang2022face2face}. For instance, Das \textit{et al.} \cite{das2020speech} employ landmarks as an intermediate representation, using an audio-to-landmark module followed by a landmark-to-image module to connect the audio inputs with video outputs. However, training these sub-modules separately can lead to error accumulation, resulting in suboptimal performance. To address this issue, alternative approaches have been developed that integrate features extracted by encoders from multiple modalities to reconstruct talking head videos in an end-to-end manner, mitigating the aforementioned limitations\cite{thies2020neural,shen2022learning,chen2018lip,Song_Zhu_Li_Wang_Qi_2019,zhou2019talking,chen2023vast,wang2023lipformer,shen2023difftalk}. Yet, these methods
still struggle to generate details that are absent from the source image. To resolve this, we introduce a memory-based approach that uses additional storage to retain these missing details for use during inference.

\section{Investigate Identity Leakage in GAN}
\label{sec:edtalk}

\subsection{Preliminaries: GAN-based methods}
\label{sec:edtalk_framework}
As shown in~\cref{fig:edtalk_limitation} (b), GAN-based models typically follow an encoder-decoder structure, involving the extraction of identity information $f^s$ from the source image and motion features $f^d$ from the driven video. To more intuitively demonstrate the limitations of GAN-based models in high-quality rendering, this paper selects the SOTA GAN-based model, EDTalk~\cite{edtalk}, as our Baseline for investigation. Unlike previous methods, this Baseline introduces a Face Decoupling Module (FDM; see \textit{suppl} for details) to disentangle facial dynamics into distinct components for decoupled control, such as mouth shapes, head poses, and emotional expressions. Once the identity feature $f^s$ and motion feature $f^d$ are extracted, they are fed into the generator $G$ to produce the final output $I^g$. To preserve the identity information, $G$ incorporates the multi-scale features $f^s_{1:5}$ extracted from the source image by encoder $E$:
\begin{equation}
\label{eq:g}
    I^g = G( f^{s}+ f^{d} , f^{s}_{1:5})
\end{equation}
Baseline is trained in a self-supervised manner to reconstruct $I^d$ using reconstruction loss $\mathcal{L}_\text{rec}$, perceptual loss $\mathcal{L}_\text{per}$~\cite{johnson2016perceptual, zhang2018unreasonable} and adversarial loss $\mathcal{L}_\text{adv}$:
\begin{equation}
\label{eq:rec_per}
   \mathcal{L}_\text{rec} = \|I^d-I^g\|_1; \qquad \mathcal{L}_\text{per} = \|\Phi(I^d)-\Phi(I^g)\|^2_2
\end{equation}
\begin{equation}
\label{eq:adv}
\mathcal{L}_\text{adv} =(\text{log}D(I^d)+\text{log}(1-D(I^g))),
\end{equation}
where $\Phi$ denotes the feature extractor of VGG19~\cite{simonyan2014very} and $D$ is a discriminator tasked with distinguishing between reconstructed images and ground truth (GT).

\subsection{Discussion about identity leakage}
\label{sec:edtalk_limitation}

Although Baseline achieves satisfactory performance, we identify some limitations. To illustrate them more clearly, we conduct tests on Baseline using: (1) In self-driven setting, we sample two frames with significant different head poses and expressions from one video as the source image $I^{s1}$ and driven image $I^{d1}$.
(2) In cross-driven setting, we keep the driven image $I^{d1}$ unchanged but use a source image $I^{s2}$ with significant differences in appearance compared to $I^{d1}$. We expect Baseline to generate satisfactory results $\hat{I}^{s2}_b, \hat{I}^{s1}_b$ in both settings—producing images with the appearance of source image and the motion of driven image.

The results are shown in~\cref{fig:edtalk_limitation}, revealing two key issues: (1) Identity Leakage: From $\hat{I}^{s2}_b$, we can observe that not only does it mimic the motion of $I^{d1}$, but its appearance, including face shape, also shifts towards $I^{d1}$. This is a prevalent problem in current methods, where the driven image's identity undesirably leaks into the source image.
The main reason is that the motion and identity features are not fully disentangled, mainly due to the lack of decoupling constraints. (2) Self-driven Result $\hat{I}^{s1}_b$ Outperforms Cross-driven Result $\hat{I}^{s2}_b$: Although both settings produce expressions and head poses as those in $I^{d1}$, and both generate noticeable artifacts due to significant differences in head pose between the source and driven images, we still find that $\hat{I}^{s1}_b$ mimics the motion $I^{d1}$ more effectively than $\hat{I}^{s2}_b$. This discrepancy arises because, while both $I^{s1}$ and $I^{d1}$ represent the same person, their poses and facial expressions are significantly different. As a result, identity information (such as appearance) of the driven image carried by motion-related features will be leaked into the source image during the self-driven process, contributing to better animation results. In other words, when the source and driven images are from the same person, the identity leakage actually aids the animation process. For better understanding, we hypothesize in extreme cases, if the driven image fully leaks its identity information into the source image, the animation process can be simplified to self reconstruction, significantly reducing artifacts and movement inaccuracy. Therefore, a question arises: Can we tame identity leakage for high-quality talking head generation in extreme poses and exaggerated expressions?

Towards answering this question, we first explore which variable contributes to the observed identity leakage. We systematically replace the intermediate variables of the driving image extracted by Baseline with the corresponding variables from the source image and generate the final results. The detailed process and complete results can be found in the supplementary material, where we conduct ablation experiments on all intermediate variables of Baseline. From our investigation, we conclude that $f^d_{1:5}$ carries the most identity leakage, as shown in~\cref{fig:edtalk_limitation} (c). By using $f^d_{1:5}$ as input to $G$ instead of $f^s_{1:5}$, we find that $\hat{I}^{s1}_c$ is able to generate details and extreme poses that $\hat{I}^{s1}_b$ cannot, while the appearance of $\hat{I}^{s2}_c$ more closely resembles that of $I^{d1}$. To further simplify the process, we identify that only the feature $f^d_{4}$ from the 4-th layer is necessary to convey the information required for identity leakage. The specific results are provided in~\cref{fig:edtalk_limitation} (d)\&(e). This finding answers our question: By utilizing the $f^d_{4}$ from the driving image, we can tame identity leakage for high-quality rendering.

\section{Method}
\label{sec:method}

\begin{figure*}[t]
  \centering
  \includegraphics[width=1\linewidth]{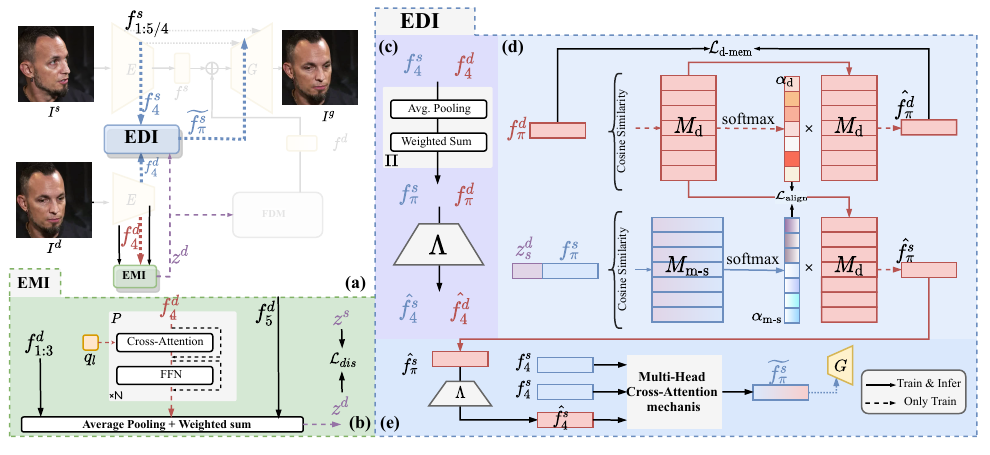}
    \caption{The illustration of FixTalk. (a) FixTalk is built upon Baseline (covered by a gray mask). (b) Unlike Baseline, which directly uses $f^d_6$ to extract motion features, FixTalk fully leverages the multi-scale features of encoder $E$, where we introduce the EMI to gain a more comprehensive understanding of motion. In EMI, a lightweight extractor $P$ autonomously learns motion-specific features from $f^d_4$. Then, the multi-scale features are passed through an Average Pooling layer and a Weighted Sum layer to generate the final motion feature $z^d$. (c)\&(d)\&(e) In contrast to Baseline which feeds $f^s_{1:5}$ into $G$, FixTalk enhances features using EDI for better detail rendering. During training, EDI employs a memory network to record the driven features. During inference, the most appropriate feature is retrieved from the memory network to supplement the details. We also support audio-driven talking head generation, whose details are contained in \textit{Suppl}.
    }
    \label{fig:FixTalk}
    \vspace{-0.1in}
\end{figure*}

Inspired by our observations in~\cref{sec:edtalk_limitation}, this paper introduces FixTalk, which inherits Baseline framework (in~\cref{sec:edtalk_framework}, masked in~\cref{fig:FixTalk} (a)), to tame identity leakage for high-quality talking heads. As shown in~\cref{fig:FixTalk}, the taming process consists of (1) avoiding identity leakage that alters the appearance of source image, for which we propose the Enhanced Motion Indicator (EMI in~\cref{sec:EMI}), and (2) harnessing identity leakage to enhance model to render extreme expressions and poses, leading to the development of the Enhanced Detail Indicator (EDI in~\cref{sec:EDI}).

\subsection{Enhanced Motion Indicator}
\label{sec:EMI}
From our analysis in~\cref{sec:edtalk_limitation}, we identify that a primary reason for identity leakage is the coupling of identity information within the extracted motion features. To address this issue, we propose the Enhanced Motion Indicator (EMI), which produces a motion feature completely disentangled from identity information by improving both the model design and the loss function. Regarding model design, we enhance the ability of encoder $E$ to understand motion variance across different scale features $f^d_{1:5}$.
As discussed in~\cref{sec:edtalk_limitation} and illustrated in~\cref{fig:edtalk_limitation} (d)\&(e), $f^d_4$ encodes not only motion information but also substantial identity information from the driven image, which is unnecessary and can be disruptive during motion extraction. To overcome this challenge, we introduce a lightweight extractor $P$, composed of $N$ stacked layers of cross-attention and feed-forward networks (FFN), as shown in~\cref{fig:FixTalk} (b). In the cross-attention layer, we use $f^d_4$ as the keys (${K}$) and values (${V}$). The main challenge lies in designing an appropriate query (${Q}$) to guide motion extraction. Inspired by the success of the Q-Former~\cite{qformer}, we initialize a learnable query vector, $q_l$, which learns during training to select the most useful features from $f^d_4$ for the subsequent motion extraction process. Next, we pass the extracted features through an Average Pooling layer to capture scale-specific information. This is followed by a Weighted Sum layer~\cite{yang2021superb} that assigns learnable weights to effectively merge information from the diverse layers, obtaining the enhanced motion feature $z^d$, which is further fed into FDM.

To completely remove identity information from the enhanced motion features, we aim to minimize the similarity between the motion features of the source image and the driven image due to their different movement. To achieve this, we introduce a disentanglement loss~\cite{emoportraits} $\mathcal{L}_\text{dis}$:
\vspace{-5pt}
\begin{equation}
\label{eq:dis}
    \mathcal{L}_\text{dis} = \text{max}(0, \text{cos}(z^s,z^d) - \xi),
\end{equation}
\vspace{-3pt}
where $z^s$ and $z^d$ are enhanced motion features extracted from $I^s$ and $I^d$ and $\xi$ adjust the strictness of the loss.

\subsection{Enhanced Detail Indicator}
\label{sec:EDI}

In EMI, we address identity leakage by isolating the identity-related features in $f^d_4$, which helps prevent appearance changes in results. Our analysis in~\cref{sec:edtalk_limitation} reveals that in the self-driven setting, the identity leakage from $f^d_4$ actually contributes to the details generation.
However, this leakage is not suitable for the cross-driven setting, and directly replacing these features can lead to severe artifacts. To tackle both issues, we propose the Enhanced Detail Indicator (EDI). First of all, to conserve space and computational resources, we introduce a feature compressor $\Pi$ composed of an average pooling layer and a weighted sum, as shown in~\cref{fig:FixTalk} (c). $\Pi$ extracts valid information from $f^{s,d}_4 \in \mathbb{R}^{512 \times 32 \times 32}$, yielding the compressed token $f^{s,d}_{\pi} \in \mathbb{R}^{1 \times 512}$. For convenient re-input into $G$, we design a decompressor $\Lambda$ to restore them back to original shape. 

Ideally, we expect that even in the cross-driven setting, the generator $G$ can benefit from $f^d_4$ associated with the same identity, enabling it to perform better in cases involving exaggerated expressions and extreme poses. Therefore, one motivation for EDI is to utilize additional storage space during training to store these features, allowing for appropriate features to be queried during inference. In self-supervised training, since the driving image $I^d$ and source image $I^s$ are sampled from the same person, the differences between $f^{s}_{\pi}$ and $f^{d}_{\pi}$ are primarily due to different motion, represented as the motion embedding difference $z^d_s = z^d - z^s$. In other words, $f^{d}_{\pi}$ can be inferred from the pair \{$z^d_s, f^{s}_{\pi}$\}. To accomplish this, we utilize a Memory Network that makes use of external memory to retain the necessary information. Specifically, in~\cref{fig:FixTalk} (d), we design a Leakage Memory Network comprising a driven-identity memory $M_{\text{d}}$ and motion-source memory $M_{\text{m-s}}$, which store the mutually aligned $f^{d}_{\pi}$ and \{$z^d_s, f^{s}_{\pi}$\}.
More precisely, the driven-identity memory $M_{\text{d}}$ = $\{m_{\text{d}}^i\}_{i=1}^S$ comprises of $S$ slots, where the $i$th slot contains the compressed driven token $m_{\text{d}}^i$. During training, the compact driven token $f^d_\pi$ extracted from $I^d$, is used as a guide for the driven-identity memory. $f^d_\pi$ is initially used as a query to compute the cosine similarity with each slot, followed by the softmax function $\Psi(\cdot)$  to produce the weight $\alpha_{\text{d}}^i$ of each slot:
\vspace{-5pt}
\begin{equation}
    \alpha_{\text{d}}^i = \Psi\left(\frac{f^d_\pi \cdot m_\text{d}^i}{\left\|f^d_\pi\right\|_2\cdot\left\|m_{\text{d}}^i\right\|_2}\right)
\end{equation}
\vspace{-5pt}
In this manner, we obtain the weights for all slots $\Omega_{\text{d}} = \left\{\alpha_{\text{d}}^1,...,\alpha_{\text{d}}^i,...,\alpha_{\text{d}}^S\right\}$. We consider $\Omega_{\text{d}}$ to serve as the address for each slot in memory when querying with $f^d_\pi$. It allows us to retrieve the feature $\hat {f^d_\pi}$ most relevant to $f^d_\pi$. To refresh the driven-identity memory, we reduce the distance between the recalled feature $\hat {f^d_\pi}$ and $f^d_\pi$:
\begin{equation}
    \hat {f^d_\pi} = \sum_{i=1}^{S}\alpha_{\text{d}}^i \cdot m_{\text{d}}^i, \quad     \mathcal{L}_{\text{d-mem}} = \left\|f^d_\pi - \hat{f^d_\pi}\right\|
\end{equation}

However, during inference, $f^d_\pi$ matching the identity of $I^s$ is not available to determine the weight of each slot. This necessitates using the accessible source feature $f^s_\pi$ and motion difference ${z^d_s}$ as keys for querying values in the driven-identity memory. Accordingly, we set up the motion-source memory $M_{\text{m-s}}$ following the same approach as for the driven-identity memory:
\begin{equation}
    \alpha_{\text{m-s}}^i = \Psi\left(\frac{(f^s_\pi\bigoplus z^d_s) \cdot m_{\text{m-s}}^i}{\left\|(f^s_\pi\bigoplus z^d_s)\right\|_2\cdot\left\|m_{\text{m-s}}^i\right\|_2}\right),
\end{equation}
where $m_{\text{m-s}}^i$ indicates the $i$th slot containing motion-source feature within the motion-source memory $M_{\text{m-s}}$, $\bigoplus$ refers to the concatenate operation. Also, the weight of each slot can be derived as $\Omega_{\text{m-s}} = \left\{\alpha_{\text{m-s}}^1,...,\alpha_{\text{m-s}}^i,...,\alpha_{\text{m-s}}^S\right\}$.

\begin{figure*}[t]
  \centering
  \includegraphics[width=1\linewidth]{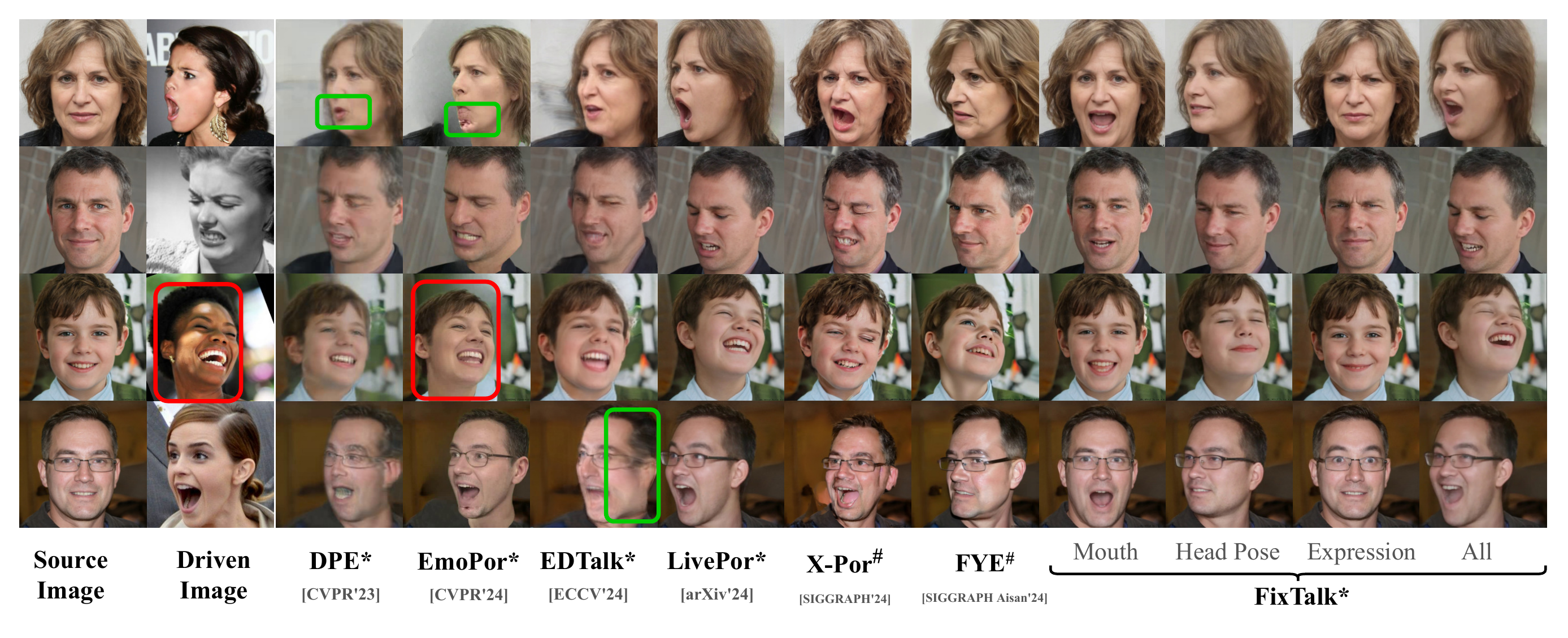}
  \vspace{-0.2in}
    \caption{Qualitative comparisons in video-driven setting. See full comparison in supplementary material (\textit{Suppl}).}
    \label{fig:video_comparison}
    \vspace{-0.2in}
\end{figure*}

To establish a correlation between driven-identity memory and motion-source memory, we need to place matching features at the corresponding addresses in both memories. To achieve this alignment, we apply the KL divergence:
\begin{equation}
    \mathcal{L}_\text{align} = KL(\Omega_{\text{m-s}} || \Omega_{\text{d}})
\end{equation}
By ensuring consistency between the key address and value address derived from  $f^s_\pi\bigoplus z^d_s$ and $f^d_\pi$, both addresses point to the same slots in the driven-identity memory, regardless of whether they are queried with $f^s_\pi\bigoplus z^d_s$ or $f^d_\pi$. Consequently, during inference, we can retrieve the most relevant compact driven token matching the identity of $I^s$ by using the available motion-source feature as the query:
\begin{equation}
    \hat{f}^s_\pi = \sum_{i=1}^{S}\alpha_{\text{m-s}}^i \cdot m_{\text{d}}^i,
\end{equation}
where $\hat{f}^s_\pi$ represents the compact driven token recalled via motion-source feature.

As shown in~\cref{fig:FixTalk} (e), once the driving feature is obtained, we use the decompressor decompressor $\Lambda$ to upscale it to $\hat{f}^s_4$. A multi-head cross-attention mechanism (MHCA)\cite{wang2022restoreformer} is applied to spatially integrate $\hat{f}^s_4$ with $f^s_4$, resulting in $\widetilde {f^s_\pi}$. This integration improves facial rendering by ensuring both detail accuracy and motion guidance. Leveraging the combined capabilities of EMI and EDI, the generator $G$ produces the final output ${I}^g$, which retains the appearance of the source image and the motion of the driven image, free of any identity leakage or artifacts. Furthermore, since FixTalk preserves the disentangled spaces defined by Baseline and both EMI and EDI are highly efficient, FixTalk maintains its initial performance while effectively addressing the limitations discussed in~\cref{sec:edtalk_limitation}. Moreover, we follow Baseline to introduce an Audio-to-Motion module, allowing for audio-driven talking head generation by simply predicting the weights of disentangled spaces from audio, where details are in \textit{Suppl}.

\section{Experiments}
\label{sec:experiments}

\subsection{Experimental Settings}
\noindent \textbf{Datasets.}
Our method is trained on two datasets: VFHQ~\cite{vfhq} and MEAD~\cite{wang2020mead}. VFHQ contains over 16,000 high-quality clips from various interview scenarios, featuring a wide range of head poses, which provides ample material for training our memory network. The MEAD dataset, with 60 actors and actresses displaying 8 different emotions, enables the model to learn a rich variety of facial expressions. For evaluation, we partition the MEAD dataset into training and testing sets following EAMM~\cite{ji2022eamm} and introduce the external HDTF~\cite{zhang2021flow} to validate the generalization capability of our method. The videos are resized to a resolution of 512x512 and a frame rate of 25 fps.

\noindent \textbf{Evaluation Metrics.}
In addition to the commonly used metrics such as PSNR~\cite{hore2010image}, SSIM~\cite{wang2004image}, and FID~\cite{Seitzer2020FID} for measuring visual quality, M/F-LMD~\cite{chen2019hierarchical} and the confidence score of SyncNet~\cite{chung2017out} for quantifying audio-visual synchronization, and emotion accuracy for evaluating emotional expressiveness, we introduce CSIM for accessing \textbf{identity leakage}, NIQE~\cite{mittal2012making} and CPBD for accessing \textbf{rendering artifacts}. Due to space limitations, we only compare FixTalk with most recent competitive baselines and provide a comparison with all baselines in \textit{Suppl}.

\subsection{Comparison with state-of-the-art methods}
\noindent \textbf{Video-driven Setting.} 
We perform a comparative analysis with SOTA methods, including DPE*~\cite{dpe}, EmoPor*~\cite{emoportraits}, LivePor*~\cite{guo2024liveportrait}, X-For$^\#$~\cite{xie2024x}, FYE$^\#$~\cite{ma2024followyouremoji} where * and $\#$ refer to the GAN-based and diffusion-based models, respectively. 
From the \textbf{qualitative results} shown in~\cref{fig:video_comparison}, GAN-based methods commonly exhibit noticeable \textcolor{leakage}{identity leakage} and \textcolor{artifacts}{rendering artifacts}. While LivePor generates the correct motions, it suffers from temporal inconsistency and lip bonding, which is especially clear in the supplementary videos. X-Por struggles to generate accurate facial expressions and head poses, and although FYE shows improvements in head pose, it still fails to generate correct expressions. In contrast, our method not only produces satisfactory results but also allows for disentangled control over the mouth, head pose, and expression, as shown in the last four columns. This is also corroborated by the \textbf{quantitative results} shown in~\cref{tab:video_comp}, highlighting the superior ability of FixTalk. Notably, identity leakage is alleviated (higher CSIM), and the rendering quality is significantly better than GAN-based models (lower NIQE and higher CPBD).

To evaluate the performance from a human perspective, we randomly sample 10 videos generated by each method and conduct a \textbf{user study}. Specifically, we invite 20 participants and ask them to rate each video on a scale of 1-5 based on their motion consistency, identity preservation, image quality. The results are summarized in~\cref{tab:user_study}, where our method outperforms others in all aspects.

\begin{table}[t]
\setlength\tabcolsep{1pt}
  \resizebox{\linewidth}{!}{

  \begin{tabular}{@{}lcccccc@{}}
    \toprule
    Method & PSNR$\uparrow$  &  F-LMD$\downarrow$ &  FID$\downarrow$   & CSIM $\uparrow$ & NIQE $\downarrow$  & CPBD $\uparrow$  \\
    \midrule
DPE~\cite{dagan} & 26.078                     & 1.232          & 23.126          & 0.567         &42.96         & 0.183   \\
    EmoPor~\cite{emoportraits} & 26.827                      & 1.413          & 26.329          &0.493         & 29.88          & 0.178  \\
    EDTalk~\cite{edtalk}  & 26.504                     & \underline{1.111 }         & 13.172          & \underline{0.594 }        &42.41    & 0.221    \\
    LivePor~\cite{guo2024liveportrait}  & \underline{ 27.054}        & 1.119          & \underline{ 12.883}    & 0.568    & \underline{15.93 }      & 0.244         \\
    X-Por~\cite{xie2024x}    & 22.884                     & 1.498          & 46.552          & 0.505         & 28.61          &0.236           \\
FYE~\cite{ma2024followyouremoji}      & 23.441                        & 1.513          & 42.681          &0.544          &     18.36   &  \underline{0.247 }      \\
\midrule

\textbf{FixTalk}  &\textbf{27.164} & \textbf{1.093} & \textbf{12.715} & \textbf{0.613} & \textbf{13.44} & \textbf{ 0.282}  \\ 

    \bottomrule
  \end{tabular}
  }
  \caption{Video-driven comparisons on HDTF.}
  \label{tab:video_comp}
  \vspace{-0.1in}
\end{table}%

\begin{table}[t]
  \centering
  \resizebox{\linewidth}{!}{
  \begin{tabular}{@{}lcccccc@{}}
    \toprule
    Method & EmoPor~\cite{emoportraits} & EDTalk~\cite{edtalk} & X-Por~\cite{xie2024x} & FYE~\cite{ma2024followyouremoji} & FixTalk & GT \\
    \midrule
    Motion & 3.95 & \underline{4.15} & 3.22 & 3.37 & \textbf{4.21} & - \\
    Identity & 3.85 & \underline{4.13} & 3.98 & 4.01 & \textbf{4.47} & - \\
    Quality &  3.99 & 4.07 & \underline{4.11} & 4.05 & \textbf{4.19} & 4.45 \\
    \bottomrule
  \end{tabular}
  }
  \caption{User study results.}
  \label{tab:user_study}
    \vspace{-0.1in}
\end{table}

\begin{table}[t]
\setlength\tabcolsep{1pt}
  \resizebox{\linewidth}{!}{

  \begin{tabular}{@{}lcccccc@{}}
    \toprule
    Metric/Method & PSNR$\uparrow$ &  SSIM$\uparrow$ &  M-LMD $\downarrow$& F-LMD$\downarrow$ &  $\text{Acc}_\text{emo}$ $\uparrow$ & $\text{Sync}_\text{conf}\uparrow$ \\
    \midrule
    SadTalker~\cite{zhang2023sadtalker}    & 19.042          & 0.606          & 2.038          & 2.335          & 14.25         & 7.065          \\
AniTalker~\cite{liu2024anitalker}    & 19.714          & 0.614          & 1.903          & 2.277          & 15.62         & 6.638          \\
Hallo~\cite{xu2024hallo}        & 19.061          & 0.598          & 1.874          & 2.294          & 18.69         & 6.993          \\
EchoMimic~\cite{echomimic}    & 18.884          & 0.600            & 1.793          & 2.110           &17.35         & 5.190           \\
FlowVQTalker~\cite{tan2024flowvqtalker} & 21.572          & 0.709          & 1.551          & 1.304          & 61.53          & 6.901          \\
EDTalk~\cite{edtalk}       & \underline{21.628}          & \underline{0.722}          & \underline{1.537}          & \underline{1.290}           & \underline{67.32}         & \textbf{8.115} \\
\midrule
\textbf{FixTalk} & \textbf{22.382} & \textbf{0.743} & \textbf{1.314} & \textbf{1.215} & \textbf{68.25} & \underline{8.009}\\
    \bottomrule
  \end{tabular}
  }
  \caption{Audio-driven quantitative comparisons on MEAD.}
  \label{tab:audio_comp}
  \vspace{-0.1in}
\end{table}%

\begin{figure*}[t]
  \centering
  \includegraphics[width=1\linewidth]{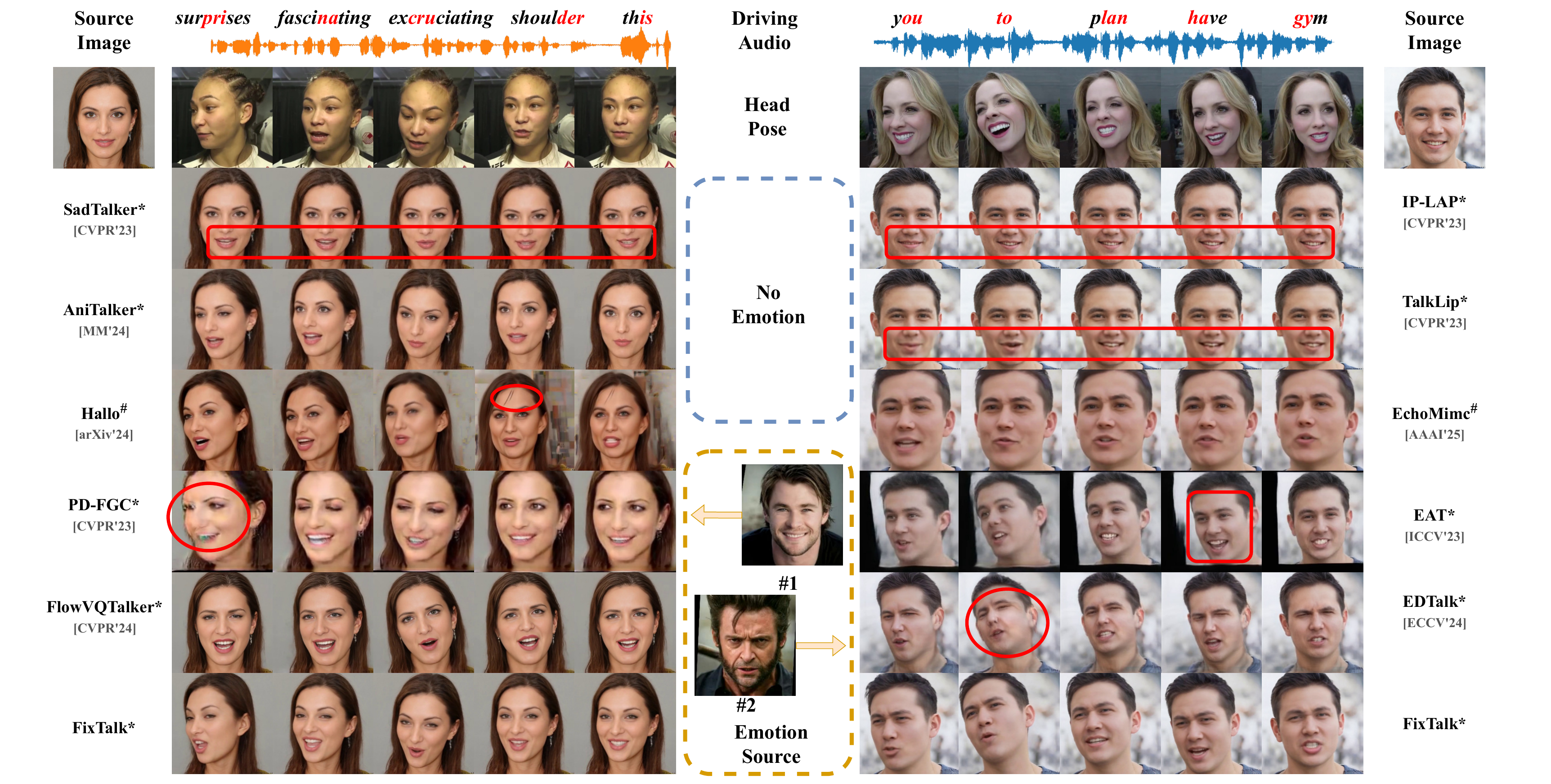}
  \vspace{-0.2in}
    \caption{Qualitative comparisons in audio-driven setting. See full comparison in supplementary material (\textit{Suppl}).}
    \label{fig:audio_comparison}
    \vspace{-0.1in}
\end{figure*}

\noindent \textbf{Audio-driven Setting.} 
We compare our method with: (a) emotion-agnostic talking face generation methods: SadTalker*~\cite{zhang2023sadtalker}, IP-LAP*~\cite{zhong2023identity}, TalkLip*~\cite{wang2023seeing}, AniTalker*~\cite{liu2024anitalker}, EchoMimic$^\#$~\cite{echomimic} and Hallo$^\#$~\cite{xu2024hallo}. (b) Emotional talking face generation methods: PD-FGC*~\cite{pd-fgc}, EAT*~\cite{gan2023efficient}, FlowVQTalker*~\cite{tan2024flowvqtalker} and EDTalk*~\cite{edtalk}. We show the keyframes in~\cref{fig:audio_comparison}. \cite{zhang2023sadtalker, zhong2023identity, wang2023seeing} are only able to generate minor lip movements. Even when we provide a pose source for \cite{liu2024anitalker}, it still fails to produce accurate head poses. \cite{xu2024hallo} suffers from a gradual decline in identity preservation during long-duration inference. Both \cite{gan2023efficient, tan2024flowvqtalker} generate incorrect emotional expressions, while \cite{pd-fgc, edtalk} exhibit artifacts when handling extreme poses. As shown in the quantitative results in~\cref{tab:audio_comp}, we achieve the best performance across most metrics, with a particularly significant improvement over EDTalk.

\begin{figure}[t]
  \centering
  \includegraphics[width=1\linewidth]{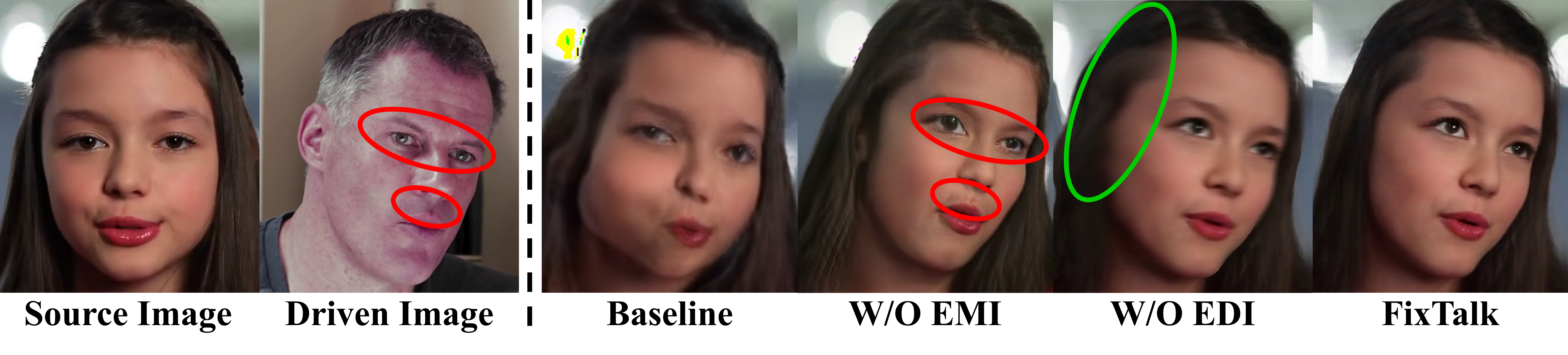}
    \caption{Ablation study results. }
    \label{fig:ablation_study}
    \vspace{-0.1in}
\end{figure}

\subsection{Ablation Study.}

In~\cref{fig:ablation_study}, we visualize the generated images after removing EMI, EDI and both (Baseline) from FixTalk. The baseline results show significant identity changes and artifacts. When only EMI is removed, the artifacts are reduced, but the character's facial features, such as face shape, remain closer to the driven image, making the appearance more masculine. On the other hand, when only EDI is removed, identity preservation is improved, but artifacts reappear. With both EMI and EDI included, FixTalk achieves better identity preservation and fewer artifacts. This demonstrates that EMI addresses identity leakage, while EDI enhances detailed rendering. Refer to \textit{Suppl} for quantitative results. 

\subsection{Discussion}
\label{sec:generalizability}
\noindent \textbf{System efficiency.}
We analyze model size and inference speed, running all models on an NVIDIA A100 (80GB) for a fair comparison. As shown in~\cref{tab:efficiency}, diffusion-based models demand high GPU memory, while our method requires only 3.65 GB and can generate 27.6 frames per second, surpassing the 25 FPS standard for real-time generation. Unlike them which rely on large-scale, privately collected high-quality datasets, we train on publicly available datasets, yet achieve competitive performance.

\begin{table}[t]
  \centering
  \resizebox{\linewidth}{!}{

  \begin{tabular}{@{}lccccc@{}}
    \toprule
    Method & Echomimic$^\#$~\cite{echomimic} & EchomimicV2$^\#$~\cite{echomimicv2} & Hallo$^\#$~\cite{xu2024hallo} & Hallo3$^\#$~\cite{cui2024hallo3} & Ours*\\
    \midrule
    $\text{Size}_\text{model}$  (GB) $\downarrow$ & 7.32      & 11.60       & 9.64  & 37.71  & \textbf{3.65} \\
    $\text{Speed}_\text{infer}$  (FPS) $\uparrow$  &0.057      & 0.281        & 0.463  & 0.075  & \textbf{27.601}  \\
    $\text{Size}_\text{data}$(H)                          & $\sim$540h                     & $\sim$160h                       & $\sim$164h                 & $\sim$134h                  & \textbf{$\sim$55h}                    \\
$\text{Public}_\text{data}$    &             \XSolidBrush                   &      \XSolidBrush                            &          \XSolidBrush                  &         \XSolidBrush                    &           \Checkmark         \\         
    \bottomrule
  \end{tabular}
  }
  \caption{Comparison of system efficiency.}
  \label{tab:efficiency}
\end{table}

\begin{figure}[t]
  \centering
  \includegraphics[width=1\linewidth]{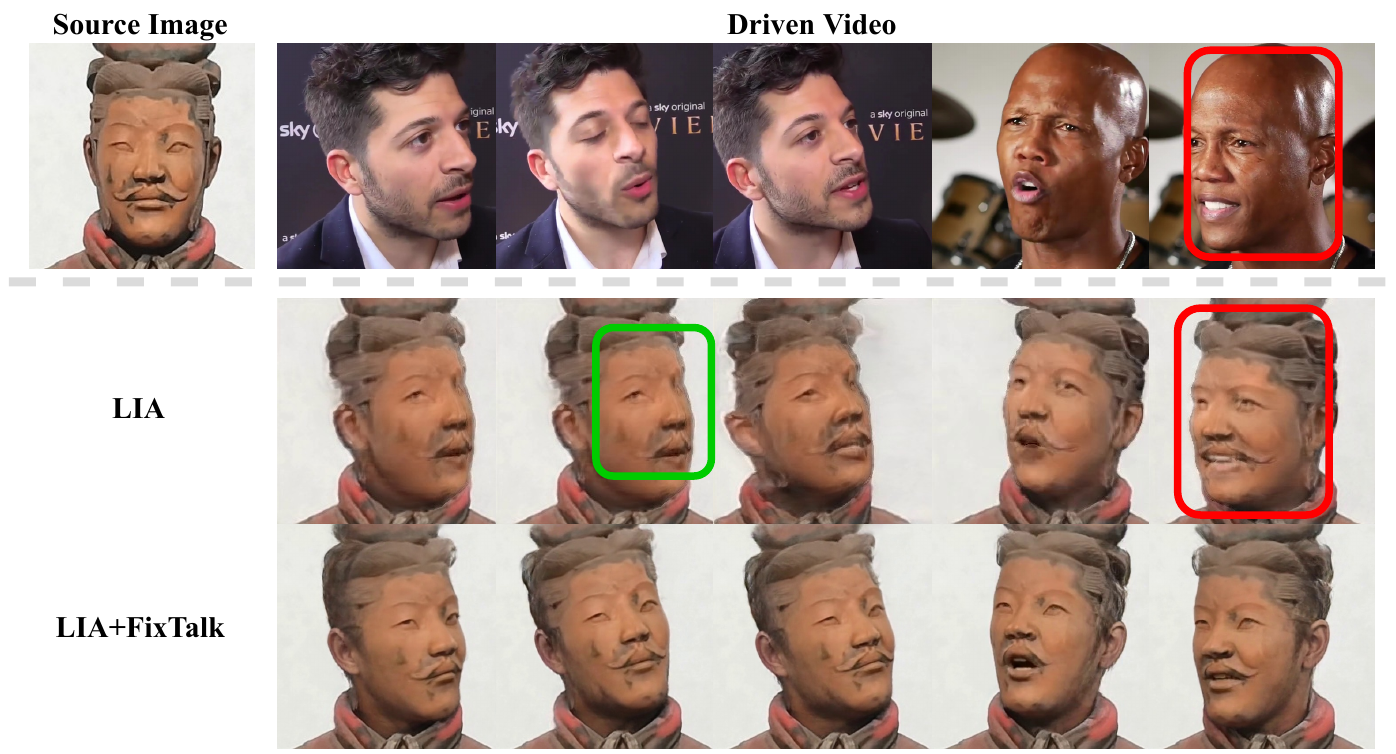}
    \caption{Comparison between LIA wo/w FixTalk.}
    \label{fig:lia++}
    \vspace{-0.1in}
\end{figure}

\noindent \textbf{Generalizability.} Our method can be seamlessly integrated into other baselines. Detailed analysis can be found in \textit{Suppl}. To verify this, we conduct experiments using LIA~\cite{lia} as baseline. As presented in~\cref{fig:lia++}, original LIA suffers from identity leakage and obvious rendering artifacts in extreme poses. In contrast, FixTalk preserves identity better and generates higher-quality results.

\noindent \textbf{Limitation.} While FixTalk has made significant progress, our method cannot infer natural emotions or subtle micro-expressions directly from audio. Moreover, while mouth, pose, and expressions are decoupled, eye movements remain linked to other facial features, limiting expressiveness.

\section{Conclusion}
This paper introduces FixTalk, a novel framework designed for generating high-quality facial animations. Building on our key findings regarding identity leakage, we propose EMI and EDI to address two prevalent issues in GAN-based models: identity leakage and rendering artifacts. FixTalk demonstrates significant improvements over SOTAs. For details on ethical considerations, please refer to the \textit{Suppl}.

{
    \small
    \bibliographystyle{ieeenat_fullname}
    \bibliography{main}

\begin{thebibliography}{99}
\providecommand{\natexlab}[1]{#1}
\providecommand{\url}[1]{\texttt{#1}}
\expandafter\ifx\csname urlstyle\endcsname\relax
  \providecommand{\doi}[1]{doi: #1}\else
  \providecommand{\doi}{doi: \begingroup \urlstyle{rm}\Url}\fi

\bibitem[Chang et~al.(2023)Chang, Shi, Gao, Fu, Xu, Song, Yan, Yang, and Soleymani]{magicdance}
Di Chang, Yichun Shi, Quankai Gao, Jessica Fu, Hongyi Xu, Guoxian Song, Qing Yan, Xiao Yang, and Mohammad Soleymani.
\newblock Magicdance: Realistic human dance video generation with motions \& facial expressions transfer.
\newblock \emph{arXiv preprint arXiv:2311.12052}, 2023.

\bibitem[Chen et~al.(2024{\natexlab{a}})Chen, Seneviratne, Wang, Hu, Saha, Hasan, Rasnayaka, Malepathirana, Gong, and Halgamuge]{chen2024anifacediff}
Ken Chen, Sachith Seneviratne, Wei Wang, Dongting Hu, Sanjay Saha, Md~Tarek Hasan, Sanka Rasnayaka, Tamasha Malepathirana, Mingming Gong, and Saman Halgamuge.
\newblock Anifacediff: High-fidelity face reenactment via facial parametric conditioned diffusion models.
\newblock \emph{arXiv preprint arXiv:2406.13272}, 2024{\natexlab{a}}.

\bibitem[Chen et~al.(2018)Chen, Li, Maddox, Duan, and Xu]{chen2018lip}
Lele Chen, Zhiheng Li, Ross~K Maddox, Zhiyao Duan, and Chenliang Xu.
\newblock Lip movements generation at a glance.
\newblock In \emph{Proceedings of the European Conference on Computer Vision (ECCV)}, pages 520--535, 2018.

\bibitem[Chen et~al.(2019)Chen, Maddox, Duan, and Xu]{chen2019hierarchical}
Lele Chen, Ross~K Maddox, Zhiyao Duan, and Chenliang Xu.
\newblock Hierarchical cross-modal talking face generation with dynamic pixel-wise loss.
\newblock In \emph{Proceedings of the IEEE/CVF conference on computer vision and pattern recognition}, pages 7832--7841, 2019.

\bibitem[Chen et~al.(2020)Chen, Cui, Liu, Li, Kou, Xu, and Xu]{chen2020talking}
Lele Chen, Guofeng Cui, Celong Liu, Zhong Li, Ziyi Kou, Yi Xu, and Chenliang Xu.
\newblock Talking-head generation with rhythmic head motion.
\newblock In \emph{European Conference on Computer Vision}, pages 35--51. Springer, 2020.

\bibitem[Chen et~al.(2023)Chen, Wu, Li, Bao, Ling, Tan, and Zhao]{chen2023vast}
Liyang Chen, Zhiyong Wu, Runnan Li, Weihong Bao, Jun Ling, Xu Tan, and Sheng Zhao.
\newblock Vast: Vivify your talking avatar via zero-shot expressive facial style transfer.
\newblock In \emph{Proceedings of the IEEE/CVF International Conference on Computer Vision}, pages 2977--2987, 2023.

\bibitem[Chen et~al.(2024{\natexlab{b}})Chen, Cao, Chen, Li, and Ma]{echomimic}
Zhiyuan Chen, Jiajiong Cao, Zhiquan Chen, Yuming Li, and Chenguang Ma.
\newblock Echomimic: Lifelike audio-driven portrait animations through editable landmark conditions.
\newblock \emph{arXiv preprint arXiv:2407.08136}, 2024{\natexlab{b}}.

\bibitem[Chung and Zisserman(2017)]{chung2017out}
Joon~Son Chung and Andrew Zisserman.
\newblock Out of time: automated lip sync in the wild.
\newblock In \emph{Computer Vision--ACCV 2016 Workshops: ACCV 2016 International Workshops, Taipei, Taiwan, November 20-24, 2016, Revised Selected Papers, Part II 13}, pages 251--263. Springer, 2017.

\bibitem[Cui et~al.(2024{\natexlab{a}})Cui, Li, Yao, Zhu, Shang, Cheng, Zhou, Zhu, and Wang]{cui2024hallo2}
Jiahao Cui, Hui Li, Yao Yao, Hao Zhu, Hanlin Shang, Kaihui Cheng, Hang Zhou, Siyu Zhu, and Jingdong Wang.
\newblock Hallo2: Long-duration and high-resolution audio-driven portrait image animation.
\newblock \emph{arXiv preprint arXiv:2410.07718}, 2024{\natexlab{a}}.

\bibitem[Cui et~al.(2024{\natexlab{b}})Cui, Li, Zhan, Shang, Cheng, Ma, Mu, Zhou, Wang, and Zhu]{cui2024hallo3}
Jiahao Cui, Hui Li, Yun Zhan, Hanlin Shang, Kaihui Cheng, Yuqi Ma, Shan Mu, Hang Zhou, Jingdong Wang, and Siyu Zhu.
\newblock Hallo3: Highly dynamic and realistic portrait image animation with diffusion transformer networks.
\newblock \emph{arXiv preprint arXiv:2412.00733}, 2024{\natexlab{b}}.

\bibitem[Das et~al.(2020)Das, Biswas, Sinha, and Bhowmick]{das2020speech}
Dipanjan Das, Sandika Biswas, Sanjana Sinha, and Brojeshwar Bhowmick.
\newblock Speech-driven facial animation using cascaded gans for learning of motion and texture.
\newblock In \emph{Computer Vision--ECCV 2020: 16th European Conference, Glasgow, UK, August 23--28, 2020, Proceedings, Part XXX 16}, pages 408--424. Springer, 2020.

\bibitem[Drobyshev et~al.(2022)Drobyshev, Chelishev, Khakhulin, Ivakhnenko, Lempitsky, and Zakharov]{megaportraits}
Nikita Drobyshev, Jenya Chelishev, Taras Khakhulin, Aleksei Ivakhnenko, Victor Lempitsky, and Egor Zakharov.
\newblock Megaportraits: One-shot megapixel neural head avatars.
\newblock In \emph{Proceedings of the 30th ACM International Conference on Multimedia}, pages 2663--2671, 2022.

\bibitem[Drobyshev et~al.(2024)Drobyshev, Casademunt, Vougioukas, Landgraf, Petridis, and Pantic]{emoportraits}
Nikita Drobyshev, Antoni~Bigata Casademunt, Konstantinos Vougioukas, Zoe Landgraf, Stavros Petridis, and Maja Pantic.
\newblock Emoportraits: Emotion-enhanced multimodal one-shot head avatars.
\newblock In \emph{Proceedings of the IEEE/CVF Conference on Computer Vision and Pattern Recognition}, pages 8498--8507, 2024.

\bibitem[Gan et~al.(2023)Gan, Yang, Yue, Sun, and Yang]{gan2023efficient}
Yuan Gan, Zongxin Yang, Xihang Yue, Lingyun Sun, and Yi Yang.
\newblock Efficient emotional adaptation for audio-driven talking-head generation.
\newblock In \emph{Proceedings of the IEEE/CVF International Conference on Computer Vision}, pages 22634--22645, 2023.

\bibitem[Goodfellow et~al.(2020)Goodfellow, Pouget-Abadie, Mirza, Xu, Warde-Farley, Ozair, Courville, and Bengio]{gan}
Ian Goodfellow, Jean Pouget-Abadie, Mehdi Mirza, Bing Xu, David Warde-Farley, Sherjil Ozair, Aaron Courville, and Yoshua Bengio.
\newblock Generative adversarial networks.
\newblock \emph{Communications of the ACM}, 63\penalty0 (11):\penalty0 139--144, 2020.

\bibitem[Guan et~al.(2023)Guan, Zhang, Zhou, Hu, Wang, He, Feng, Liu, Ding, Liu, et~al.]{guan2023stylesync}
Jiazhi Guan, Zhanwang Zhang, Hang Zhou, Tianshu Hu, Kaisiyuan Wang, Dongliang He, Haocheng Feng, Jingtuo Liu, Errui Ding, Ziwei Liu, et~al.
\newblock Stylesync: High-fidelity generalized and personalized lip sync in style-based generator.
\newblock In \emph{Proceedings of the IEEE/CVF Conference on Computer Vision and Pattern Recognition}, pages 1505--1515, 2023.

\bibitem[Guan et~al.(2024)Guan, Xu, Zhou, Wang, He, Zhang, Liang, Feng, Ding, Liu, et~al.]{resyncer}
Jiazhi Guan, Zhiliang Xu, Hang Zhou, Kaisiyuan Wang, Shengyi He, Zhanwang Zhang, Borong Liang, Haocheng Feng, Errui Ding, Jingtuo Liu, et~al.
\newblock Resyncer: Rewiring style-based generator for unified audio-visually synced facial performer.
\newblock \emph{arXiv preprint arXiv:2408.03284}, 2024.

\bibitem[Guo et~al.(2024)Guo, Zhang, Liu, Zhong, Zhang, Wan, and Zhang]{guo2024liveportrait}
Jianzhu Guo, Dingyun Zhang, Xiaoqiang Liu, Zhizhou Zhong, Yuan Zhang, Pengfei Wan, and Di Zhang.
\newblock Liveportrait: Efficient portrait animation with stitching and retargeting control.
\newblock \emph{arXiv preprint arXiv:2407.03168}, 2024.

\bibitem[Han et~al.(2024)Han, Zhu, Feng, Ji, He, Li, Liu, et~al.]{mimaface}
Yue Han, Junwei Zhu, Yuxiang Feng, Xiaozhong Ji, Keke He, Xiangtai Li, Yong Liu, et~al.
\newblock Mimaface: Face animation via motion-identity modulated appearance feature learning.
\newblock \emph{arXiv preprint arXiv:2409.15179}, 2024.

\bibitem[Hong et~al.(2022)Hong, Zhang, Shen, and Xu]{dagan}
Fa-Ting Hong, Longhao Zhang, Li Shen, and Dan Xu.
\newblock Depth-aware generative adversarial network for talking head video generation.
\newblock In \emph{Proceedings of the IEEE/CVF conference on computer vision and pattern recognition}, pages 3397--3406, 2022.

\bibitem[Hore and Ziou(2010)]{hore2010image}
Alain Hore and Djemel Ziou.
\newblock Image quality metrics: Psnr vs. ssim.
\newblock In \emph{2010 20th international conference on pattern recognition}, pages 2366--2369. IEEE, 2010.

\bibitem[Hu et~al.(2023)Hu, Gao, Zhang, Sun, Zhang, and Bo]{Animateanyone}
Li Hu, Xin Gao, Peng Zhang, Ke Sun, Bang Zhang, and Liefeng Bo.
\newblock Animate anyone: Consistent and controllable image-to-video synthesis for character animation.
\newblock \emph{arXiv preprint arXiv:2311.17117}, 2023.

\bibitem[Jamaludin et~al.(2019)Jamaludin, Chung, and Zisserman]{jamaludin2019you}
Amir Jamaludin, Joon~Son Chung, and Andrew Zisserman.
\newblock You said that?: Synthesising talking faces from audio.
\newblock \emph{International Journal of Computer Vision}, 127:\penalty0 1767--1779, 2019.

\bibitem[Ji et~al.(2023)Ji, Pan, Yan, Chen, and Yang]{ji2023stylevr}
Bin Ji, Ye Pan, Yichao Yan, Ruizhao Chen, and Xiaokang Yang.
\newblock Stylevr: Stylizing character animations with normalizing flows.
\newblock \emph{IEEE Transactions on Visualization and Computer Graphics}, 2023.

\bibitem[Ji et~al.(2025)Ji, Pan, Liu, Tan, and Yang]{ji2025sport}
Bin Ji, Ye Pan, Zhimeng Liu, Shuai Tan, and Xiaokang Yang.
\newblock Sport: From zero-shot prompts to real-time motion generation.
\newblock \emph{IEEE Transactions on Visualization and Computer Graphics}, 2025.

\bibitem[Ji et~al.(2022)Ji, Zhou, Wang, Wu, Wu, Xu, and Cao]{ji2022eamm}
Xinya Ji, Hang Zhou, Kaisiyuan Wang, Qianyi Wu, Wayne Wu, Feng Xu, and Xun Cao.
\newblock Eamm: One-shot emotional talking face via audio-based emotion-aware motion model.
\newblock In \emph{ACM SIGGRAPH 2022 Conference Proceedings}, pages 1--10, 2022.

\bibitem[Jiang et~al.(2024)Jiang, Liang, Yang, Lin, Zhong, and Zheng]{loopy}
Jianwen Jiang, Chao Liang, Jiaqi Yang, Gaojie Lin, Tianyun Zhong, and Yanbo Zheng.
\newblock Loopy: Taming audio-driven portrait avatar with long-term motion dependency.
\newblock \emph{arXiv preprint arXiv:2409.02634}, 2024.

\bibitem[Johnson et~al.(2016)Johnson, Alahi, and Fei-Fei]{johnson2016perceptual}
Justin Johnson, Alexandre Alahi, and Li Fei-Fei.
\newblock Perceptual losses for real-time style transfer and super-resolution.
\newblock In \emph{Computer Vision--ECCV 2016: 14th European Conference, Amsterdam, The Netherlands, October 11-14, 2016, Proceedings, Part II 14}, pages 694--711. Springer, 2016.

\bibitem[Li et~al.(2023)Li, Li, Savarese, and Hoi]{qformer}
Junnan Li, Dongxu Li, Silvio Savarese, and Steven Hoi.
\newblock Blip-2: Bootstrapping language-image pre-training with frozen image encoders and large language models.
\newblock In \emph{International conference on machine learning}, pages 19730--19742. PMLR, 2023.

\bibitem[Liang et~al.(2024)Liang, Zhang, Li, Yu, and Xu]{liang2024intergen}
Han Liang, Wenqian Zhang, Wenxuan Li, Jingyi Yu, and Lan Xu.
\newblock Intergen: Diffusion-based multi-human motion generation under complex interactions.
\newblock \emph{International Journal of Computer Vision}, 132\penalty0 (9):\penalty0 3463--3483, 2024.

\bibitem[Lin et~al.(2024)Lin, Jiang, Liang, Zhong, Yang, and Zheng]{lin2024cyberhost}
Gaojie Lin, Jianwen Jiang, Chao Liang, Tianyun Zhong, Jiaqi Yang, and Yanbo Zheng.
\newblock Cyberhost: Taming audio-driven avatar diffusion model with region codebook attention.
\newblock \emph{arXiv preprint arXiv:2409.01876}, 2024.

\bibitem[Liu et~al.(2024{\natexlab{a}})Liu, Chen, Fan, Du, Chen, Chen, and Yu]{liu2024anitalker}
Tao Liu, Feilong Chen, Shuai Fan, Chenpeng Du, Qi Chen, Xie Chen, and Kai Yu.
\newblock Anitalker: Animate vivid and diverse talking faces through identity-decoupled facial motion encoding.
\newblock \emph{arXiv preprint arXiv:2405.03121}, 2024{\natexlab{a}}.

\bibitem[Liu et~al.(2024{\natexlab{b}})Liu, Ma, Chen, Chen, Fan, Chen, and Yu]{liu2024vqtalker}
Tao Liu, Ziyang Ma, Qi Chen, Feilong Chen, Shuai Fan, Xie Chen, and Kai Yu.
\newblock Vqtalker: Towards multilingual talking avatars through facial motion tokenization.
\newblock \emph{arXiv preprint arXiv:2412.09892}, 2024{\natexlab{b}}.

\bibitem[Ma et~al.(2024{\natexlab{a}})Ma, Zhang, Sun, Yan, Han, and Xie]{ma2024cvthead}
Haoyu Ma, Tong Zhang, Shanlin Sun, Xiangyi Yan, Kun Han, and Xiaohui Xie.
\newblock Cvthead: One-shot controllable head avatar with vertex-feature transformer.
\newblock In \emph{Proceedings of the IEEE/CVF Winter Conference on Applications of Computer Vision}, pages 6131--6141, 2024{\natexlab{a}}.

\bibitem[Ma et~al.(2024{\natexlab{b}})Ma, Liu, Wang, Pan, He, Yuan, Zeng, Cai, Shum, Liu, et~al.]{ma2024followyouremoji}
Yue Ma, Hongyu Liu, Hongfa Wang, Heng Pan, Yingqing He, Junkun Yuan, Ailing Zeng, Chengfei Cai, Heung-Yeung Shum, Wei Liu, et~al.
\newblock Follow-your-emoji: Fine-controllable and expressive freestyle portrait animation.
\newblock \emph{arXiv preprint arXiv:2406.01900}, 2024{\natexlab{b}}.

\bibitem[Meng et~al.(2024)Meng, Zhang, Li, and Ma]{echomimicv2}
Rang Meng, Xingyu Zhang, Yuming Li, and Chenguang Ma.
\newblock Echomimicv2: Towards striking, simplified, and semi-body human animation.
\newblock \emph{arXiv preprint arXiv:2411.10061}, 2024.

\bibitem[Mittal et~al.(2012)Mittal, Soundararajan, and Bovik]{mittal2012making}
Anish Mittal, Rajiv Soundararajan, and Alan~C Bovik.
\newblock Making a “completely blind” image quality analyzer.
\newblock \emph{IEEE Signal processing letters}, 20\penalty0 (3):\penalty0 209--212, 2012.

\bibitem[Pan et~al.(2023)Pan, Zhang, Cheng, Tan, Ding, Mitchell, and Yang]{pan2023emotional}
Ye Pan, Ruisi Zhang, Shengran Cheng, Shuai Tan, Yu Ding, Kenny Mitchell, and Xubo Yang.
\newblock Emotional voice puppetry.
\newblock \emph{IEEE Transactions on Visualization and Computer Graphics}, 29\penalty0 (5):\penalty0 2527--2535, 2023.

\bibitem[Pan et~al.(2024)Pan, Tan, Cheng, Lin, Zeng, and Mitchell]{pan2024expressive}
Ye Pan, Shuai Tan, Shengran Cheng, Qunfen Lin, Zijiao Zeng, and Kenny Mitchell.
\newblock Expressive talking avatars.
\newblock \emph{IEEE Transactions on Visualization and Computer Graphics}, 2024.

\bibitem[Pan et~al.(2025)Pan, Liu, Xu, Tan, and Yang]{pan2025vasa}
Ye Pan, Chang Liu, Sicheng Xu, Shuai Tan, and Jiaolong Yang.
\newblock Vasa-rig: Audio-driven 3d facial animation with ‘live’mood dynamics in virtual reality.
\newblock \emph{IEEE Transactions on Visualization and Computer Graphics}, 2025.

\bibitem[Pang et~al.(2023)Pang, Zhang, Quan, Fan, Cun, Shan, and Yan]{dpe}
Youxin Pang, Yong Zhang, Weize Quan, Yanbo Fan, Xiaodong Cun, Ying Shan, and Dong-ming Yan.
\newblock Dpe: Disentanglement of pose and expression for general video portrait editing.
\newblock In \emph{Proceedings of the IEEE/CVF Conference on Computer Vision and Pattern Recognition}, pages 427--436, 2023.

\bibitem[Peng et~al.(2024)Peng, Wang, Zhang, Li, Yang, and Jia]{peng2024controlnext}
Bohao Peng, Jian Wang, Yuechen Zhang, Wenbo Li, Ming-Chang Yang, and Jiaya Jia.
\newblock Controlnext: Powerful and efficient control for image and video generation.
\newblock \emph{arXiv preprint arXiv:2408.06070}, 2024.

\bibitem[Rombach et~al.(2022)Rombach, Blattmann, Lorenz, Esser, and Ommer]{stablediffusion}
Robin Rombach, Andreas Blattmann, Dominik Lorenz, Patrick Esser, and Bj{\"o}rn Ommer.
\newblock High-resolution image synthesis with latent diffusion models.
\newblock In \emph{CVPR}, pages 10684--10695, 2022.

\bibitem[Seitzer(2020)]{Seitzer2020FID}
Maximilian Seitzer.
\newblock {pytorch-fid: FID Score for PyTorch}.
\newblock \url{https://github.com/mseitzer/pytorch-fid}, 2020.
\newblock Version 0.3.0.

\bibitem[Shen et~al.(2022)Shen, Li, Zhu, Duan, Zhou, and Lu]{shen2022learning}
Shuai Shen, Wanhua Li, Zheng Zhu, Yueqi Duan, Jie Zhou, and Jiwen Lu.
\newblock Learning dynamic facial radiance fields for few-shot talking head synthesis.
\newblock In \emph{European Conference on Computer Vision}, pages 666--682. Springer, 2022.

\bibitem[Shen et~al.(2023)Shen, Zhao, Meng, Li, Zhu, Zhou, and Lu]{shen2023difftalk}
Shuai Shen, Wenliang Zhao, Zibin Meng, Wanhua Li, Zheng Zhu, Jie Zhou, and Jiwen Lu.
\newblock Difftalk: Crafting diffusion models for generalized audio-driven portraits animation.
\newblock In \emph{Proceedings of the IEEE/CVF Conference on Computer Vision and Pattern Recognition}, pages 1982--1991, 2023.

\bibitem[Shi et~al.(2025)Shi, Gong, Chen, Zheng, Tan, Yang, Li, He, Zheng, Chen, et~al.]{shi2024motionstone}
Shuwei Shi, Biao Gong, Xi Chen, Dandan Zheng, Shuai Tan, Zizheng Yang, Yuyuan Li, Jingwen He, Kecheng Zheng, Jingdong Chen, et~al.
\newblock Motionstone: Decoupled motion intensity modulation with diffusion transformer for image-to-video generation.
\newblock \emph{CVPR 2025}, 2025.

\bibitem[Simonyan and Zisserman(2014)]{simonyan2014very}
Karen Simonyan and Andrew Zisserman.
\newblock Very deep convolutional networks for large-scale image recognition.
\newblock \emph{arXiv preprint arXiv:1409.1556}, 2014.

\bibitem[Song et~al.(2019)Song, Zhu, Li, Wang, and Qi]{Song_Zhu_Li_Wang_Qi_2019}
Yang Song, Jingwen Zhu, Dawei Li, Andy Wang, and Hairong Qi.
\newblock Talking face generation by conditional recurrent adversarial network.
\newblock In \emph{Proceedings of the Twenty-Eighth International Joint Conference on Artificial Intelligence}, 2019.

\bibitem[Tan et~al.(2023)Tan, Ji, and Pan]{tan2023emmn}
Shuai Tan, Bin Ji, and Ye Pan.
\newblock Emmn: Emotional motion memory network for audio-driven emotional talking face generation.
\newblock In \emph{Proceedings of the IEEE/CVF International Conference on Computer Vision}, pages 22146--22156, 2023.

\bibitem[Tan et~al.(2024{\natexlab{a}})Tan, Ji, Ding, and Pan]{tan2024say}
Shuai Tan, Bin Ji, Yu Ding, and Ye Pan.
\newblock Say anything with any style.
\newblock In \emph{Proceedings of the AAAI Conference on Artificial Intelligence}, pages 5088--5096, 2024{\natexlab{a}}.

\bibitem[Tan et~al.(2024{\natexlab{b}})Tan, Ji, and Pan]{tan2024flowvqtalker}
Shuai Tan, Bin Ji, and Ye Pan.
\newblock Flowvqtalker: High-quality emotional talking face generation through normalizing flow and quantization.
\newblock In \emph{Proceedings of the IEEE/CVF Conference on Computer Vision and Pattern Recognition}, pages 26317--26327, 2024{\natexlab{b}}.

\bibitem[Tan et~al.(2024{\natexlab{c}})Tan, Ji, and Pan]{tan2024style2talker}
Shuai Tan, Bin Ji, and Ye Pan.
\newblock Style2talker: High-resolution talking head generation with emotion style and art style.
\newblock In \emph{Proceedings of the AAAI Conference on Artificial Intelligence}, pages 5079--5087, 2024{\natexlab{c}}.

\bibitem[Tan et~al.(2025{\natexlab{a}})Tan, Gong, Feng, Zheng, Zheng, Shi, Shen, Chen, and Yang]{mimir}
Shuai Tan, Biao Gong, Yutong Feng, Kecheng Zheng, Dandan Zheng, Shuwei Shi, Yujun Shen, Jingdong Chen, and Ming Yang.
\newblock Mimir: Improving video diffusion models for precise text understanding.
\newblock In \emph{Proceedings of the Computer Vision and Pattern Recognition Conference}, pages 23978--23988, 2025{\natexlab{a}}.

\bibitem[Tan et~al.(2025{\natexlab{b}})Tan, Gong, Wang, Zhang, Zheng, Zheng, Zheng, Chen, and Yang]{tan2024animate}
Shuai Tan, Biao Gong, Xiang Wang, Shiwei Zhang, Dandan Zheng, Ruobing Zheng, Kecheng Zheng, Jingdong Chen, and Ming Yang.
\newblock Animate-x: Universal character image animation with enhanced motion representation.
\newblock In \emph{ICLR 2025}, 2025{\natexlab{b}}.

\bibitem[Tan et~al.(2025{\natexlab{c}})Tan, Gong, Wei, Zhang, Liu, Zheng, Chen, Wang, Ouyang, Zheng, and Shen]{tan2025SynMotion}
Shuai Tan, Biao Gong, Yujie Wei, Shiwei Zhang, Zhuoxin Liu, Dandan Zheng, Jingdong Chen, Yan Wang, Hao Ouyang, Kecheng Zheng, and Yujun Shen.
\newblock Synmotion: Semantic-visual adaptation for motion customized video generation.
\newblock \emph{arXiv preprint arXiv:2506.23690}, 2025{\natexlab{c}}.

\bibitem[Tan et~al.(2025{\natexlab{d}})Tan, Ji, Bi, and Pan]{edtalk}
Shuai Tan, Bin Ji, Mengxiao Bi, and Ye Pan.
\newblock Edtalk: Efficient disentanglement for emotional talking head synthesis.
\newblock In \emph{European Conference on Computer Vision}, pages 398--416. Springer, 2025{\natexlab{d}}.

\bibitem[Tan et~al.(2024{\natexlab{d}})Tan, Lin, Xu, Ji, Zhu, Wang, and Fu]{tan2024svp}
Weipeng Tan, Chuming Lin, Chengming Xu, Xiaozhong Ji, Junwei Zhu, Chengjie Wang, and Yanwei Fu.
\newblock Svp: Style-enhanced vivid portrait talking head diffusion model.
\newblock \emph{arXiv preprint arXiv:2409.03270}, 2024{\natexlab{d}}.

\bibitem[Thies et~al.(2020)Thies, Elgharib, Tewari, Theobalt, and Nie{\ss}ner]{thies2020neural}
Justus Thies, Mohamed Elgharib, Ayush Tewari, Christian Theobalt, and Matthias Nie{\ss}ner.
\newblock Neural voice puppetry: Audio-driven facial reenactment.
\newblock In \emph{Computer Vision--ECCV 2020: 16th European Conference, Glasgow, UK, August 23--28, 2020, Proceedings, Part XVI 16}, pages 716--731. Springer, 2020.

\bibitem[Tian et~al.(2024)Tian, Wang, Zhang, and Bo]{emo}
Linrui Tian, Qi Wang, Bang Zhang, and Liefeng Bo.
\newblock Emo: Emote portrait alive-generating expressive portrait videos with audio2video diffusion model under weak conditions.
\newblock \emph{arXiv preprint arXiv:2402.17485}, 2024.

\bibitem[Tong et~al.(2024)Tong, Li, Chen, Wu, and Zhou]{musepose}
Zhengyan Tong, Chao Li, Zhaokang Chen, Bin Wu, and Wenjiang Zhou.
\newblock Musepose: a pose-driven image-to-video framework for virtual human generation.
\newblock \emph{arxiv}, 2024.

\bibitem[Tran and Liu(2018)]{tran2018nonlinear}
Luan Tran and Xiaoming Liu.
\newblock Nonlinear 3d face morphable model.
\newblock In \emph{Proceedings of the IEEE conference on computer vision and pattern recognition}, pages 7346--7355, 2018.

\bibitem[Wang et~al.(2024{\natexlab{a}})Wang, Tian, Zhang, Guan, Luo, Shen, Jiang, Gu, Han, and Yang]{vexpress}
Cong Wang, Kuan Tian, Jun Zhang, Yonghang Guan, Feng Luo, Fei Shen, Zhiwei Jiang, Qing Gu, Xiao Han, and Wei Yang.
\newblock V-express: Conditional dropout for progressive training of portrait video generation.
\newblock \emph{arXiv preprint arXiv:2406.02511}, 2024{\natexlab{a}}.

\bibitem[Wang et~al.(2023{\natexlab{a}})Wang, Deng, Yin, Shum, and Wang]{pd-fgc}
Duomin Wang, Yu Deng, Zixin Yin, Heung-Yeung Shum, and Baoyuan Wang.
\newblock Progressive disentangled representation learning for fine-grained controllable talking head synthesis.
\newblock In \emph{Proceedings of the IEEE/CVF Conference on Computer Vision and Pattern Recognition}, pages 17979--17989, 2023{\natexlab{a}}.

\bibitem[Wang et~al.(2023{\natexlab{b}})Wang, Qian, Zhang, Tan, and Li]{wang2023seeing}
Jiadong Wang, Xinyuan Qian, Malu Zhang, Robby~T Tan, and Haizhou Li.
\newblock Seeing what you said: Talking face generation guided by a lip reading expert.
\newblock In \emph{Proceedings of the IEEE/CVF Conference on Computer Vision and Pattern Recognition}, pages 14653--14662, 2023{\natexlab{b}}.

\bibitem[Wang et~al.(2023{\natexlab{c}})Wang, Zhao, Zhang, Zhang, Shen, Zhao, and Zhou]{wang2023lipformer}
Jiayu Wang, Kang Zhao, Shiwei Zhang, Yingya Zhang, Yujun Shen, Deli Zhao, and Jingren Zhou.
\newblock Lipformer: High-fidelity and generalizable talking face generation with a pre-learned facial codebook.
\newblock In \emph{Proceedings of the IEEE/CVF Conference on Computer Vision and Pattern Recognition}, pages 13844--13853, 2023{\natexlab{c}}.

\bibitem[Wang et~al.(2020)Wang, Wu, Song, Yang, Wu, Qian, He, Qiao, and Loy]{wang2020mead}
Kaisiyuan Wang, Qianyi Wu, Linsen Song, Zhuoqian Yang, Wayne Wu, Chen Qian, Ran He, Yu Qiao, and Chen~Change Loy.
\newblock Mead: A large-scale audio-visual dataset for emotional talking-face generation.
\newblock In \emph{Computer Vision--ECCV 2020: 16th European Conference, Glasgow, UK, August 23--28, 2020, Proceedings, Part XXI}, pages 700--717. Springer, 2020.

\bibitem[Wang et~al.(2021{\natexlab{a}})Wang, Li, Ding, Fan, and Yu]{wang2021audio2head}
S Wang, L Li, Y Ding, C Fan, and X Yu.
\newblock Audio2head: Audio-driven one-shot talking-head generation with natural head motion.
\newblock In \emph{International Joint Conference on Artificial Intelligence}. IJCAI, 2021{\natexlab{a}}.

\bibitem[Wang et~al.(2024{\natexlab{b}})Wang, Li, Lin, Lin, Yang, Zhang, Liu, and Wang]{disco}
Tan Wang, Linjie Li, Kevin Lin, Chung-Ching Lin, Zhengyuan Yang, Hanwang Zhang, Zicheng Liu, and Lijuan Wang.
\newblock Disco: Disentangled control for referring human dance generation in real world.
\newblock In \emph{ICLR}, 2024{\natexlab{b}}.

\bibitem[Wang et~al.(2024{\natexlab{c}})Wang, Zhang, Gao, Wang, Zhou, Zhang, Yan, and Sang]{wang2024unianimate}
Xiang Wang, Shiwei Zhang, Changxin Gao, Jiayu Wang, Xiaoqiang Zhou, Yingya Zhang, Luxin Yan, and Nong Sang.
\newblock Unianimate: Taming unified video diffusion models for consistent human image animation.
\newblock \emph{arXiv preprint arXiv:2406.01188}, 2024{\natexlab{c}}.

\bibitem[Wang et~al.(2021{\natexlab{b}})Wang, Yang, Bremond, and Dantcheva]{lia}
Yaohui Wang, Di Yang, Francois Bremond, and Antitza Dantcheva.
\newblock Latent image animator: Learning to animate images via latent space navigation.
\newblock In \emph{International Conference on Learning Representations}, 2021{\natexlab{b}}.

\bibitem[Wang et~al.(2004)Wang, Bovik, Sheikh, and Simoncelli]{wang2004image}
Zhou Wang, Alan~C Bovik, Hamid~R Sheikh, and Eero~P Simoncelli.
\newblock Image quality assessment: from error visibility to structural similarity.
\newblock \emph{IEEE Transactions on Image Processing}, 13\penalty0 (4):\penalty0 600--612, 2004.

\bibitem[Wang et~al.(2022)Wang, Zhang, Chen, Wang, and Luo]{wang2022restoreformer}
Zhouxia Wang, Jiawei Zhang, Runjian Chen, Wenping Wang, and Ping Luo.
\newblock Restoreformer: High-quality blind face restoration from undegraded key-value pairs.
\newblock In \emph{Proceedings of the IEEE/CVF conference on computer vision and pattern recognition}, pages 17512--17521, 2022.

\bibitem[Wei et~al.(2024)Wei, Yang, and Wang]{aniportrait}
Huawei Wei, Zejun Yang, and Zhisheng Wang.
\newblock Aniportrait: Audio-driven synthesis of photorealistic portrait animation.
\newblock \emph{arXiv preprint arXiv:2403.17694}, 2024.

\bibitem[Wu et~al.(2021)Wu, Zhang, Wu, Wang, Li, Sun, and Li]{wu2021f3a}
Xintian Wu, Qihang Zhang, Yiming Wu, Huanyu Wang, Songyuan Li, Lingyun Sun, and Xi Li.
\newblock F$^3$a-gan: Facial flow for face animation with generative adversarial networks.
\newblock \emph{IEEE Transactions on Image Processing}, 30:\penalty0 8658--8670, 2021.

\bibitem[Xie et~al.(2022)Xie, Wang, Zhang, Dong, and Shan]{vfhq}
Liangbin Xie, Xintao Wang, Honglun Zhang, Chao Dong, and Ying Shan.
\newblock Vfhq: A high-quality dataset and benchmark for video face super-resolution.
\newblock In \emph{The IEEE Conference on Computer Vision and Pattern Recognition Workshops (CVPRW)}, 2022.

\bibitem[Xie et~al.(2024)Xie, Xu, Song, Wang, Shi, and Luo]{xie2024x}
You Xie, Hongyi Xu, Guoxian Song, Chao Wang, Yichun Shi, and Linjie Luo.
\newblock X-portrait: Expressive portrait animation with hierarchical motion attention.
\newblock In \emph{ACM SIGGRAPH 2024 Conference Papers}, pages 1--11, 2024.

\bibitem[Xu et~al.(2024{\natexlab{a}})Xu, Li, Su, Shang, Zhang, Liu, Wang, Van~Gool, Yao, and Zhu]{xu2024hallo}
Mingwang Xu, Hui Li, Qingkun Su, Hanlin Shang, Liwei Zhang, Ce Liu, Jingdong Wang, Luc Van~Gool, Yao Yao, and Siyu Zhu.
\newblock Hallo: Hierarchical audio-driven visual synthesis for portrait image animation.
\newblock \emph{arXiv preprint arXiv:2406.08801}, 2024{\natexlab{a}}.

\bibitem[Xu et~al.(2024{\natexlab{b}})Xu, Chen, Guo, Yang, Li, Zang, Zhang, Tong, and Guo]{xu2024vasa}
Sicheng Xu, Guojun Chen, Yu-Xiao Guo, Jiaolong Yang, Chong Li, Zhenyu Zang, Yizhong Zhang, Xin Tong, and Baining Guo.
\newblock Vasa-1: Lifelike audio-driven talking faces generated in real time.
\newblock \emph{arXiv preprint arXiv:2404.10667}, 2024{\natexlab{b}}.

\bibitem[Xu et~al.(2023)Xu, Zhang, Liew, Yan, Liu, Zhang, Feng, and Shou]{magicanimate}
Zhongcong Xu, Jianfeng Zhang, Jun~Hao Liew, Hanshu Yan, Jia-Wei Liu, Chenxu Zhang, Jiashi Feng, and Mike~Zheng Shou.
\newblock Magicanimate: Temporally consistent human image animation using diffusion model.
\newblock \emph{arXiv preprint arXiv:2311.16498}, 2023.

\bibitem[Yang et~al.(2024{\natexlab{a}})Yang, Li, Wu, Li, Wang, Yu, Su, and Xu]{yang2024smgdiff}
Hongdi Yang, Chengyang Li, Zhenxuan Wu, Gaozheng Li, Jingya Wang, Jingyi Yu, Zhuo Su, and Lan Xu.
\newblock Smgdiff: Soccer motion generation using diffusion probabilistic models.
\newblock \emph{arXiv preprint arXiv:2411.16216}, 2024{\natexlab{a}}.

\bibitem[Yang et~al.(2022)Yang, Chen, Guo, Zhang, Guo, and Zhang]{yang2022face2face}
Kewei Yang, Kang Chen, Daoliang Guo, Song-Hai Zhang, Yuan-Chen Guo, and Weidong Zhang.
\newblock Face2face $\rho$: Real-time high-resolution one-shot face reenactment.
\newblock In \emph{European conference on computer vision}, pages 55--71. Springer, 2022.

\bibitem[Yang et~al.(2024{\natexlab{b}})Yang, Li, Wu, Jing, Li, Ji, Liang, and Fan]{yang2024megactor}
Shurong Yang, Huadong Li, Juhao Wu, Minhao Jing, Linze Li, Renhe Ji, Jiajun Liang, and Haoqiang Fan.
\newblock Megactor: Harness the power of raw video for vivid portrait animation.
\newblock \emph{arXiv preprint arXiv:2405.20851}, 2024{\natexlab{b}}.

\bibitem[Yang et~al.(2021)Yang, Chi, Chuang, Lai, Lakhotia, Lin, Liu, Shi, Chang, Lin, et~al.]{yang2021superb}
Shu-wen Yang, Po-Han Chi, Yung-Sung Chuang, Cheng-I~Jeff Lai, Kushal Lakhotia, Yist~Y Lin, Andy~T Liu, Jiatong Shi, Xuankai Chang, Guan-Ting Lin, et~al.
\newblock Superb: Speech processing universal performance benchmark.
\newblock \emph{arXiv preprint arXiv:2105.01051}, 2021.

\bibitem[Yao et~al.(2020)Yao, Yuan, Shao, and Zhou]{yao2020mesh}
Guangming Yao, Yi Yuan, Tianjia Shao, and Kun Zhou.
\newblock Mesh guided one-shot face reenactment using graph convolutional networks.
\newblock In \emph{Proceedings of the 28th ACM international conference on multimedia}, pages 1773--1781, 2020.

\bibitem[Zakharov et~al.(2019)Zakharov, Shysheya, Burkov, and Lempitsky]{zakharov2019few}
Egor Zakharov, Aliaksandra Shysheya, Egor Burkov, and Victor Lempitsky.
\newblock Few-shot adversarial learning of realistic neural talking head models.
\newblock In \emph{Proceedings of the IEEE/CVF international conference on computer vision}, pages 9459--9468, 2019.

\bibitem[Zhang et~al.(2023{\natexlab{a}})Zhang, Qi, Zhang, Zhang, Wu, Chen, Chen, Wang, and Wen]{zhang2023metaportrait}
Bowen Zhang, Chenyang Qi, Pan Zhang, Bo Zhang, HsiangTao Wu, Dong Chen, Qifeng Chen, Yong Wang, and Fang Wen.
\newblock Metaportrait: Identity-preserving talking head generation with fast personalized adaptation.
\newblock In \emph{Proceedings of the IEEE/CVF Conference on Computer Vision and Pattern Recognition}, pages 22096--22105, 2023{\natexlab{a}}.

\bibitem[Zhang et~al.(2018)Zhang, Isola, Efros, Shechtman, and Wang]{zhang2018unreasonable}
Richard Zhang, Phillip Isola, Alexei~A Efros, Eli Shechtman, and Oliver Wang.
\newblock The unreasonable effectiveness of deep features as a perceptual metric.
\newblock In \emph{Proceedings of the IEEE conference on computer vision and pattern recognition}, pages 586--595, 2018.

\bibitem[Zhang et~al.(2023{\natexlab{b}})Zhang, Cun, Wang, Zhang, Shen, Guo, Shan, and Wang]{zhang2023sadtalker}
Wenxuan Zhang, Xiaodong Cun, Xuan Wang, Yong Zhang, Xi Shen, Yu Guo, Ying Shan, and Fei Wang.
\newblock Sadtalker: Learning realistic 3d motion coefficients for stylized audio-driven single image talking face animation.
\newblock In \emph{Proceedings of the IEEE/CVF Conference on Computer Vision and Pattern Recognition}, pages 8652--8661, 2023{\natexlab{b}}.

\bibitem[Zhang et~al.(2024)Zhang, Gu, Wang, Wang, Cheng, Zhu, and Zou]{mimicmotion2024}
Yuang Zhang, Jiaxi Gu, Li-Wen Wang, Han Wang, Junqi Cheng, Yuefeng Zhu, and Fangyuan Zou.
\newblock Mimicmotion: High-quality human motion video generation with confidence-aware pose guidance.
\newblock \emph{arXiv preprint arXiv:2406.19680}, 2024.

\bibitem[Zhang et~al.(2021)Zhang, Li, Ding, and Fan]{zhang2021flow}
Zhimeng Zhang, Lincheng Li, Yu Ding, and Changjie Fan.
\newblock Flow-guided one-shot talking face generation with a high-resolution audio-visual dataset.
\newblock In \emph{Proceedings of the IEEE/CVF Conference on Computer Vision and Pattern Recognition}, pages 3661--3670, 2021.

\bibitem[Zhao and Zhang(2022)]{zhao2022thin}
Jian Zhao and Hui Zhang.
\newblock Thin-plate spline motion model for image animation.
\newblock In \emph{Proceedings of the IEEE/CVF Conference on Computer Vision and Pattern Recognition}, pages 3657--3666, 2022.

\bibitem[Zhao et~al.(2024)Zhao, Long, Zhang, Qin, Liang, Zhang, Zhang, Yu, and Xu]{zhao2024media2face}
Qingcheng Zhao, Pengyu Long, Qixuan Zhang, Dafei Qin, Han Liang, Longwen Zhang, Yingliang Zhang, Jingyi Yu, and Lan Xu.
\newblock Media2face: Co-speech facial animation generation with multi-modality guidance.
\newblock In \emph{ACM SIGGRAPH 2024 conference papers}, pages 1--13, 2024.

\bibitem[Zhao et~al.(2021)Zhao, Wu, and Guo]{zhao2021sparse}
Ruiqi Zhao, Tianyi Wu, and Guodong Guo.
\newblock Sparse to dense motion transfer for face image animation.
\newblock In \emph{Proceedings of the IEEE/CVF International Conference on Computer Vision}, pages 1991--2000, 2021.

\bibitem[Zhong et~al.(2023)Zhong, Fang, Cai, Wei, Zhao, Lin, and Li]{zhong2023identity}
Weizhi Zhong, Chaowei Fang, Yinqi Cai, Pengxu Wei, Gangming Zhao, Liang Lin, and Guanbin Li.
\newblock Identity-preserving talking face generation with landmark and appearance priors.
\newblock In \emph{Proceedings of the IEEE/CVF Conference on Computer Vision and Pattern Recognition}, pages 9729--9738, 2023.

\bibitem[Zhou et~al.(2019)Zhou, Liu, Liu, Luo, and Wang]{zhou2019talking}
Hang Zhou, Yu Liu, Ziwei Liu, Ping Luo, and Xiaogang Wang.
\newblock Talking face generation by adversarially disentangled audio-visual representation.
\newblock In \emph{Proceedings of the AAAI conference on artificial intelligence}, pages 9299--9306, 2019.

\bibitem[Zhou et~al.(2020)Zhou, Han, Shechtman, Echevarria, Kalogerakis, and Li]{zhou2020makelttalk}
Yang Zhou, Xintong Han, Eli Shechtman, Jose Echevarria, Evangelos Kalogerakis, and Dingzeyu Li.
\newblock Makelttalk: speaker-aware talking-head animation.
\newblock \emph{ACM Transactions On Graphics (TOG)}, 39\penalty0 (6):\penalty0 1--15, 2020.

\bibitem[Zhu et~al.(2024)Zhu, Chen, Dai, Xu, Cao, Yao, Zhu, and Zhu]{zhu2024champ}
Shenhao Zhu, Junming~Leo Chen, Zuozhuo Dai, Yinghui Xu, Xun Cao, Yao Yao, Hao Zhu, and Siyu Zhu.
\newblock Champ: Controllable and consistent human image animation with 3d parametric guidance.
\newblock In \emph{European Conference on Computer Vision (ECCV)}, 2024.

\bibitem[Zhu et~al.(2017)Zhu, Liu, Lei, and Li]{zhu2017face}
Xiangyu Zhu, Xiaoming Liu, Zhen Lei, and Stan~Z Li.
\newblock Face alignment in full pose range: A 3d total solution.
\newblock \emph{IEEE transactions on pattern analysis and machine intelligence}, 41\penalty0 (1):\penalty0 78--92, 2017.

\end{thebibliography}
}


\end{document}